\documentclass{article}

    \PassOptionsToPackage{numbers, compress}{natbib}

\usepackage[final]{neurips_data_2024}
\usepackage{array}
\usepackage{adjustbox}
\usepackage{multirow}
\usepackage{tabularx}
\usepackage{graphicx}
\usepackage{algorithm}
\usepackage{xcolor}
\usepackage{algorithmic}
\usepackage{multicol}
\usepackage{amsmath}
\usepackage{lipsum}
\usepackage{wrapfig}
\usepackage{makecell}
\usepackage{longtable}
\usepackage{hyperref}
\usepackage{enumitem}
\usepackage{lipsum}
\usepackage{titletoc}
\usepackage{geometry}

\usepackage[utf8]{inputenc} 
\usepackage[T1]{fontenc}    
\usepackage{hyperref}       
\usepackage{url}            
\usepackage{booktabs}       
\usepackage{amsfonts}       
\usepackage{nicefrac}       
\usepackage{microtype}      
\usepackage{xcolor}         
\usepackage{xspace}
\usepackage{fontawesome5}
\usepackage[defaultcolor=magenta]{changes}
\newcommand{\dname}{\texttt{EHRCon}\xspace}
\newcommand{\mname}{CheckEHR\xspace}
\usepackage[most]{tcolorbox}
\usepackage{tabularray}
\newcommand{\equal}[1]{{\hypersetup{linkcolor=black}\thanks{#1}}}

\title{EHRCon: Dataset for Checking Consistency between Unstructured Notes and Structured Tables in Electronic Health Records}

\author{
    Yeonsu Kwon$^{1}$\equal{These authors contributed equally}\;,
    Jiho Kim$^{1}$\footnotemark[1]\;,
    Gyubok Lee$^{1}$, 
    Seongsu Bae$^{1}$,
    Daeun Kyung$^{1}$,\\
    \textbf{Wonchul Cha$^{2}$},
    \textbf{Tom Pollard$^{3}$}, 
    \textbf{Alistair Johnson$^{4}$},  
    \textbf{Edward Choi$^{1}$} \\
    $^{1}$KAIST \;
    $^{2}$Samsung Medical Center \;
    $^{3}$MIT \;
    $^{4}$University of Toronto\\
    \texttt{\{yeonsu.k, jiho.kim, edwardchoi\}@kaist.ac.kr}}

\begin{document}

\maketitle
\begin{abstract}
  
 Electronic Health Records (EHRs) are integral for storing comprehensive patient medical records, combining structured data (\textit{e.g.}, medications) with detailed clinical notes (\textit{e.g.}, physician notes). 
 These elements are essential for straightforward data retrieval and provide deep, contextual insights into patient care. However, they often suffer from discrepancies due to unintuitive EHR system designs and human errors, posing serious risks to patient safety.
 To address this, we developed \dname, a new dataset and task specifically designed to ensure data consistency between structured tables and unstructured notes in EHRs.
 \dname was crafted in collaboration with healthcare professionals using the MIMIC-III EHR dataset, and includes manual annotations of 4,101 entities across 105 clinical notes checked against database entries for consistency.
\dname has two versions, one using the original MIMIC-III schema, and another using the OMOP CDM schema, in order to increase its applicability and generalizability.
Furthermore, leveraging the capabilities of large language models, we introduce \mname, a novel framework for verifying the consistency between clinical notes and database tables. \mname utilizes an eight-stage process and shows promising results in both few-shot and zero-shot settings. The code is available at \url{https://github.com/dustn1259/EHRCon}.
\end{abstract}

\section{Introduction}
Electronic Health Records (EHRs) are digital datasets comprising the rich information of a patient's medical history within hospitals.
These records integrate both structured data (\textit{e.g}., medications, diagnoses) and detailed clinical notes (\textit{e.g}., physician notes).
The structured data facilitates straightforward retrieval and analysis of essential information, while clinical notes provide in-depth, contextual insights into the patient's condition. 
These two forms of data are interconnected and provide complementary information throughout the diagnostic and treatment processes. For example, a practitioner might start by reviewing test results stored in the database, then determine a diagnosis and formulate a treatment plan, which are documented in the clinical notes. These notes are subsequently used to update the structured data in the database.

However, inconsistencies can arise between the two sets of data for several reasons. One primary issue is that EHR interfaces are often designed with a focus on administrative and financial tasks, which makes it difficult to accurately document clinical information~\cite{villa2014review}. 
Additionally, overburdened practitioners might unintentionally introduce errors by importing incorrect medication lists, copying and pasting outdated records, or entering inaccurate test results~\cite{bell2020frequency,payne2018using,yadav2017comparison}. 
These errors can lead to significant discrepancies between the structured data and clinical notes in the EHR, potentially jeopardizing patient safety and leading to legal complications~\cite{aziz2015electronic}.

\begin{figure}[t]
    \begin{center}
    \includegraphics[width=\linewidth]{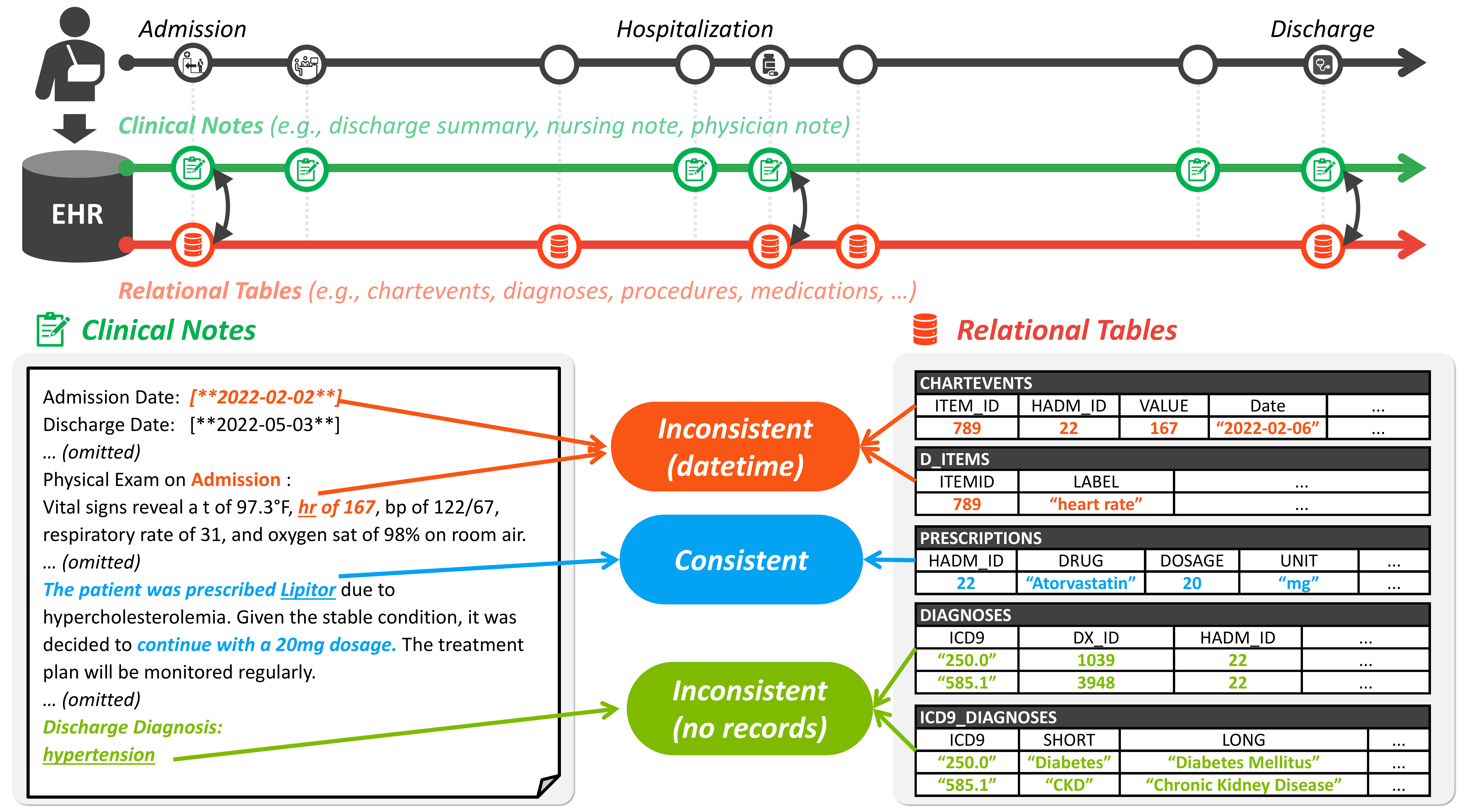}
    \vspace{-7mm} 
    \end{center}
    \caption{Examples of consistent and inconsistent data between clinical notes and EHR tables: An inconsistent example (datetime) is when a clinical note records an HR (abbreviation for heart rate) of 167 on ``2022-02-02'' but the EHR table shows the same HR on ``2022-02-06''. A consistent example is when both the clinical note and the EHR table document the administration of Atorvastatin with matching drug name, dosage, and unit. Another example of inconsistency occurs when a clinical note mentions a hypertension diagnosis, but the EHR table lacks this information.}
    \vspace{-3mm} 
    \label{figure1}
\end{figure}

Manual scrutiny of these records is both time-intensive and costly, underscoring the necessity for automated interventions.
Despite the need for automated systems, previous studies on consistency check between tables and text have primarily focused on single claims and small-scale single tables~\citep{aly2021feverous,chen2019tabfact, gupta-etal-2020-infotabs, wang-etal-2021-semeval}.
These approaches are not designed for the complex and large-scale nature of EHRs, which require more comprehensive and scalable solutions.

To this end, we propose a new task and dataset called \dname, which is designed to verify the consistency between clinical notes and large-scale relational databases in EHRs. 
We collaborated closely with practitioners\footnote{EHR technician, nurse, and emergency medicine specialist with over 15 years of experience} to design labeling instructions based on their insights and expertise, authentically reflecting real hospital environments.
Based on these labeling instructions, trained human annotators used the MIMIC-III EHR dataset~\cite{johnson2016physionet} to manually compare 4,101 entities mentioned in 105 clinical notes against corresponding table contents, annotating them for \textsc{Consistent} or \textsc{Inconsistent} as illustrated in Figure \ref{figure1}. 
Our dataset also offers interpretability by including detailed information about the specific tables and columns where inconsistencies occurred.
Moreover, it contains two versions, one based on the original MIMIC-III schema, and another based on its OMOP CDM~\cite{omop} implementation, allowing us to incorporate various schema types and enhance the generalizability.

Additionally, we introduce \mname, a framework that leverages the reasoning capabilities of large language models (LLMs) to verify consistency between clinical notes and tables in EHRs.
\mname comprises eight sequential stages, enabling it to address complex tasks in both few-shot and zero-shot settings.
Experimental results indicate that in a few-shot setting, our framework achieves a recall performance of 61.06\% on MIMIC-III and 54.36\% on OMOP.
In a zero-shot setting, it achieves a recall performance of 52.39\% on MIMIC-III.
Additionally, we conduct comprehensive ablation studies to thoroughly analyze the contributions of each component within \mname.

\label{intro}

\section{Related Works}
\textbf{Consistency Check}~~Fact verification involves assessing the truthfulness of claims by comparing them to evidence~\cite{jiang2020hover, kim-etal-2023-factkg, park2021faviq, rani-etal-2023-factify, saakyan2021covid, sarrouti2021evidence, schuster2021get, schuster2019towards, thorne2018fever, vladika2023scientific}. This task is similar to ours as it involves checking for consistency between two sets of data. Among the various datasets, those utilizing tables as evidence are particularly relevant.
TabFact~\cite{chen2019tabfact}, a prominent dataset for table-based fact verification, focuses on verifying claims by reasoning with Wikipedia tables. Additionally, INFOTABS~\cite{gupta-etal-2020-infotabs} uses info-boxes from Wikipedia, and SEM-TAB-FACTS~\cite{wang-etal-2021-semeval} utilizes tables from scientific articles. Furthermore, FEVEROUS~\cite{aly2021feverous} is a dataset designed to verify claims by reasoning over both text and tables from Wikipedia.
While these datasets focus on verifying individual claims with small-scale tables (\textit{e.g.}, most 50 rows), our methodology differs significantly.
We handle entire clinical notes where multiple claims must be first recognized, then perform consistency checks against a larger heterogeneous relational database (\textit{i.e.}, 13 tables each with up to 330M rows).
This requires a more comprehensive and scalable solution for fact verification.
Consequently, our work extends beyond previous studies, presenting a novel task in the field of Natural Language Processing (NLP) as well as healthcare.

\textbf{Compositional Reasoning}~~Large Language Models (LLMs)~\cite{anil2023palm, brown2020language, openai2023gpt4, chatgpt} have demonstrated remarkable abilities in handling a wide range of tasks with just a few examples in the prompts (\textit{i.e.}, in-context learning). However, some complex tasks remain challenging when tackled through in-context learning alone. To address these challenges, researchers have developed methods to break down complex problems into smaller, more manageable sub-tasks~\cite{khot2022decomposed,lu2024chameleon, zhou2022least}. These decomposition techniques have also been applied to tasks that involve reasoning from structured data~\cite{jiang-etal-2023-structgpt, kim-etal-2023-kg,wang2023chain}. One significant development is StructGPT~\cite{jiang-etal-2023-structgpt}, which enables LLMs to gather evidence and reason using structured data to answer questions. 
Inspired by these decomposition techniques, \mname improves the accuracy and efficiency of consistency checks between clinical notes and tables, effectively overcoming the limitations of in-context learning methods.

\section{EHRCon}
\dname includes annotations for 4,101 entities extracted from 105 randomly selected clinical notes, evaluated against 13 tables within the MIMIC-III database~\cite{johnson2016physionet}.\footnote{Although MIMIC-IV~\cite{johnson2020mimic} is more recent than MIMIC-III, we use MIMIC-III in this work because MIMIC-IV lacks diverse note types (such as physician notes and nursing notes), and is missing all dates in notes for de-identification, as opposed to shifting the dates in MIMIC-III notes.} MIMIC-III contains data from approximately 40,000 ICU patients treated at Beth Israel Deaconess Medical Center between 2001 and 2012, encompassing both structured information and textual records.
To enhance standardization, we also utilize the Observational Medical Outcomes Partnership (OMOP) Common Data Model (CDM)\footnote{https://www.ohdsi.org/data-standardization/} version of MIMIC-III. 
OMOP CDM, a publicly developed data standard designed to unify the format and content of observational data for efficient and reliable biomedical research.  In this regard, developing the OMOP version of \dname will be highly beneficial for future research scalability.
In this section, we detail the process of designing the labeling instructions (Sec.~\ref{sec3.1}), and labeling the dataset (Sec.~\ref{sec3.3}) on MIMIC-III dataset. The labeling for the OMOP CDM version and detailed data preparation steps are provided in Appendix \ref{app:mimic and omop}.

\subsection{Labeling Instructions}
\label{sec3.1}

To reflect actual hospital environments, practitioners and AI researchers collaboratively designed the labeling instructions.
The following are the three important aspects of the labeling instruction, and more detailed instructions can be found in Appendix \ref{app: labeling_instruction_rules}.

\textbf{Labels}~~
We classify the entities as either \textsc{Consistent} or \textsc{Inconsistent} based on their alignment with the tables.
This approach is fundamentally different from traditional fact-checking methods, which typically determine whether claims are \textsc{Supported} or \textsc{Refuted} using texts or tables as definitive evidence. 
In contrast, in the context of EHR, both tables and clinical notes can contain errors, making it impossible to define one as definitive evidence.
Therefore, a more flexible approach is used by labeling them as \textsc{Consistent} or \textsc{Inconsistent}.
An entity is labeled as \textsc{Consistent} if all related information, such as values and dates in the note, matches exactly with the tables.
Conversely, if even one value differs, it is labeled as \textsc{Inconsistent}.

\textbf{Definition of Entity Types}~~We categorized the entities in the notes into two main types for labeling. 
First, entities with numerical values, such as ``\textit{WBC 10.0}'', are defined as Type 1.
Second, entities without values but whose existence can be verified in the database, such as ``\textit{Vancomycin was started.}'', are labeled as Type 2. 
In our study, we did not label entities with string values because they can be represented in various ways within a database.
For example, the phrase ``\textit{BP was stable}.'' might be shown in a \textit{value} column as ``\textit{Stable}'' or ``\textit{Normal}'', or it might be indicated by numeric values in the database.
This variability can lead to labeling errors.
However, to support future research, we included them as Type 3 entities in our dataset, but did not use them in the main experiments.

\textbf{Time Expression}~~Clinical notes contain various time expressions, so we manually analyzed the time expressions in these notes (see Appendix \ref{app: time_expression}). As a result, we found that they can be categorized into three groups: 1) event time written in a standard time format, 2) event time described in a narrative style, and 3) time information of the event not written. When an entity is presented in the standard date and time format (\textit{i.e.}, \textit{YYYY-MM-DD}, \textit{YYYY-MM-DD HH:MM:SS}), we validate whether a clinical event occurred exactly at that timestamp.
For narrative-style expressions, such as ``\textit{around the time of patient admission}'' or ``\textit{shortly after discharge}'', we consider records within the day before and after the specified date to account for the approximate nature of the timing.
In cases where no precise time information is provided, we determine the relevant time frame based on the type of the note.
For instance, in discharge summaries, we examine the entire admission period, while for physician and nursing notes, we check the records within one day before and after the chart date.
\begin{figure}[t]
    \includegraphics[width=\linewidth]{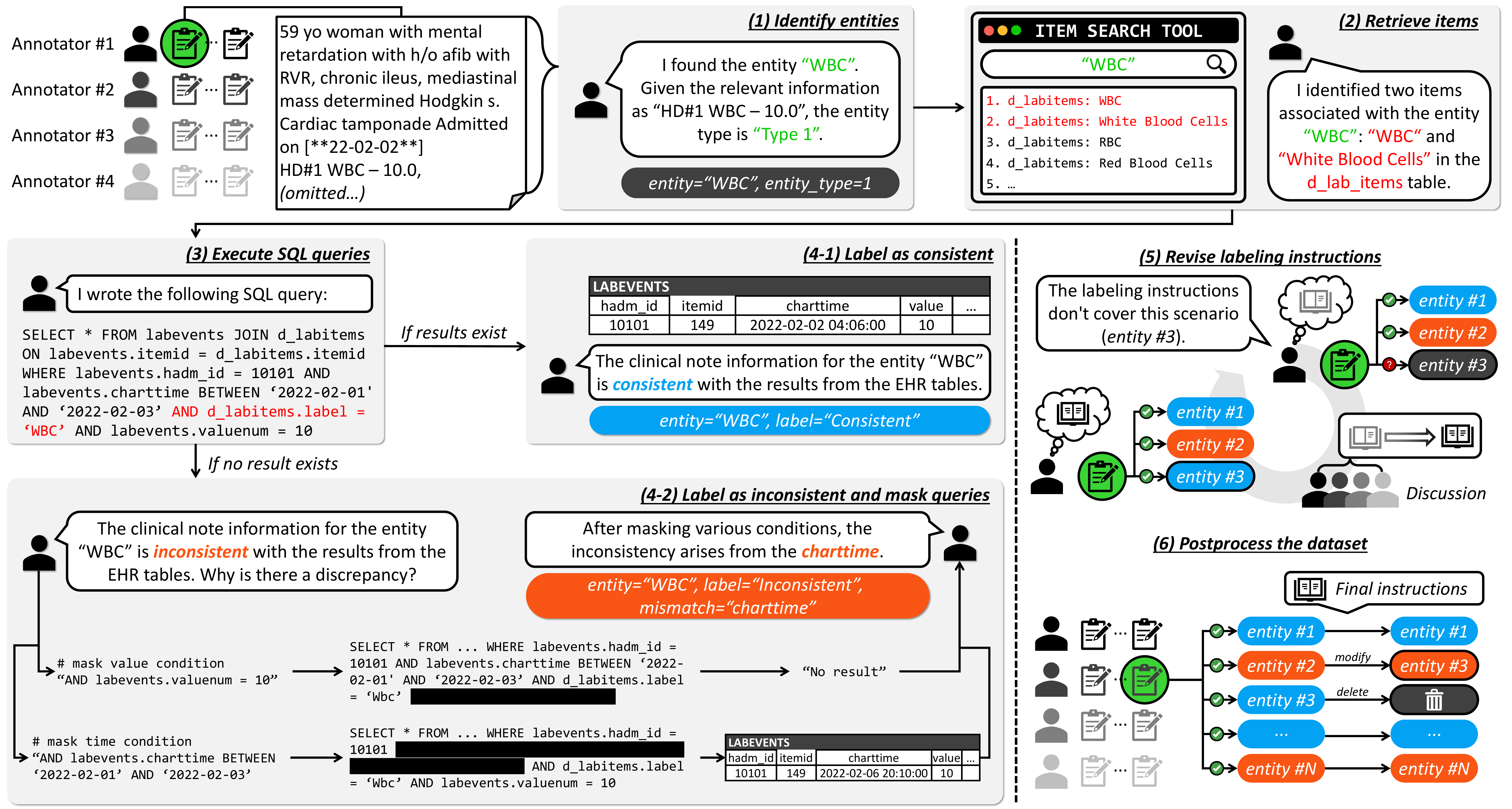}
    \caption{Annotation process of \dname: The annotation process involves annotators reviewing clinical notes, identifying and classifying entities into Type 1 and Type 2, and extracting relevant information to generate and execute SQL queries. If the SQL queries yield no results, conditions (\textit{e.g.}, value or time) are masked to pinpoint where the inconsistency occurred. When annotators encounter corner cases, they update the labeling instructions through discussion. After all labeling is complete, a post-processing phase is conducted to ensure high-quality data.}
    \label{figure2}
\end{figure}
\subsection{Item Search Tool}
\label{sec3.2}
Clinical notes can include a mix of abbreviations (\textit{e.g.}, \textit{Temp} vs. \textit{Temperature}), common names (\textit{e.g.}, \textit{White Count} vs. \textit{White Blood Cells}), and brand names (\textit{e.g.}, \textit{Tylenol} vs. \textit{Acetaminophen}) depending on the context and the practitioner's preference.
This discrepancy causes issues where the entities noted in the clinical notes do not match exactly with the items in the database. 
To resolve this, we developed a tool to search for database items related to the note entities. 

To create a set $E$ of database items related to the entity $e$, we followed a detailed approach. 
First, we used the C4-WSRS medical abbreviation dataset~\cite{c4wsrs} to gather a thorough list of abbreviation-full name pairs. 
Then, we utilized GPT-4 (0613)~\cite{openai2023gpt4}\footnote{In all cases where GPTs were used, the HIPAA-compliant GPT models provided by Azure were used.} to extract medication brand names from clinical notes and convert them to their generic names. 
By combining these methods, we built an extensive set $V$, which includes abbreviations, full names, brand names, and generic names associated with the entity $e$. Finally, to create the set $E$, we calculated the bi-gram cosine similarity scores between the elements in $V$ and the items in our database, retrieving those that exceeded a specific threshold.

\renewcommand{\arraystretch}{1.1}
\begin{table}[]
\centering
\caption{Data statistics of \dname.}
\resizebox{\textwidth}{!}{%
\begin{tabular}{l|ccc|c|ccl}
\Xhline{1.5pt}
\multirow{2}{*}{\textbf{Note Type}} & \multicolumn{3}{c|}{\textbf{Entity}} & \textbf{Labels} & \multicolumn{3}{c}{\textbf{Note}} \\ \cline{2-8} 
 & \textbf{Mean Num} & \textbf{Total Num} & \textbf{Type 1 / 2} & \textbf{Con. / Incon.} & \textbf{Total Num} & \multicolumn{2}{c}{\textbf{Mean Length}} \\ \hline
Discharge Summary & 50.21 & 1,908 & 1,400 / 508 & 1,181 / 727 & 38 & \multicolumn{2}{c}{2,789} \\ \hline
Physician Note & 46.36 & 1,530 & 1,111 / 419 & 1,230 / 300 & 33 & \multicolumn{2}{c}{1,859} \\ \hline
Nursing Note & 19.50 & 663 & \;500 / 163 & 522 / 141 & 34 & \multicolumn{2}{c}{1,111} \\ \hline
Total & 39.06 & 4,101 & 3,011 / 1,090 & 2,933 / 1,168 & 105 & \multicolumn{2}{c}{1,953} \\ \Xhline{1.5pt}
\end{tabular}%
}
\label{tab_stat}
\vspace{-4mm}
\end{table}

\subsection{Annotation Process}

\label{sec3.3}
In this section, we explain the data annotation process depicted in Figure \ref{figure2}. 
Annotators begin by carefully reviewing the clinical notes, utilizing web searches and discussions with GPT-4 (0613).
Through this process, they identify entities and relevant information within the notes.
Subsequently, the identified entities are classified into Type 1 and Type 2, as outlined in Sec.~\ref{sec3.1} (Figure~\ref{figure2}-(1)).
For each entity, annotators use the Item Search Tool (Sec.~\ref{sec3.2}) to find the relevant items in the database (Figure~\ref{figure2}-(2)).
They then select the items and tables associated with the entity. 
If none of the retrieved items match the entity, the annotators manually find and match the appropriate items.
Following this, the annotators extract information related to the entity from the notes (\textit{e.g.}, dates, values, units) and use them to generate SQL queries, as explained in Appendix~\ref{sql_gen_methodology}  (Figure~\ref{figure2}-(3)). 
Finally, the annotators execute the generated queries and review the results to label the entity as either \textsc{Consistent} or \textsc{Inconsistent} (Figure~\ref{figure2}-(4)). 
If a query yields no results, the SQL conditions are sequentially masked and executed to pinpoint the source of the inconsistency (Figure~\ref{figure2}-(4)-2).
Also, when the annotators encounter a corner case that is not addressed in the existing instructions, they update the instructions after thorough discussion (Figure~\ref{figure2}-(5)).
Upon completing all annotations, the annotators engaged in a post-processing phase to ensure high-quality data.
This phase involved additional annotation of entities according to the final labeling instructions, as well as the removal of any misaligned entities (Figure~\ref{figure2}-(6)).
We implemented additional quality control processes to ensure high-quality. For more details on these processes, see Appendix \ref{app:quality control}.

\subsection{Statistics}
\textbf{Inconsistencies Found in Notes}~~As seen in Table~\ref{tab_stat},
discharge summaries account for a significant portion of inconsistent cases, with 727 out of 1,168 total cases. Unlike nursing and physician notes, which document clinical events as they occur, discharge summaries are written at the time of discharge and summarize major events and treatments. This timing could potentially increase the likelihood of errors.
Given the pivotal role discharge summaries play in hospitals such as during inpatient-outpatient transitions \cite{black2017transitions}, inconsistencies in these notes can negatively impact patient care.

\textbf{Inconsistencies Found in Tables}~~We found 36.55\% of inconsistencies in the labevents table and 17.58\% in medication-related tables (\textit{e.g}., prescriptions). These data are crucial for patient care, and such discrepancies can lead to misdiagnosis and inaccurate medication administration, potentially resulting in patient death \cite{hammerling2012review,palchuk2010unintended}. 
Therefore, implementing automated consistency checks is important to ensure data accuracy and consistency.

\textbf{Inconsistencies Found in Columns}~~An in-depth analysis revealed that 56.16\% of discrepancies are related to time, with 58.23\% of these temporal inconsistencies involving a one-hour difference between tables and clinical notes, possibly due to issues in the EHR system \cite{ward2015effects}. This suggests that the discrepancy could result from not only human but also software issues. For a more detailed analysis, refer to Appendix \ref{app: error_cases}.

\section{\mname}
\begin{figure}[t]
    \includegraphics[width=\linewidth]{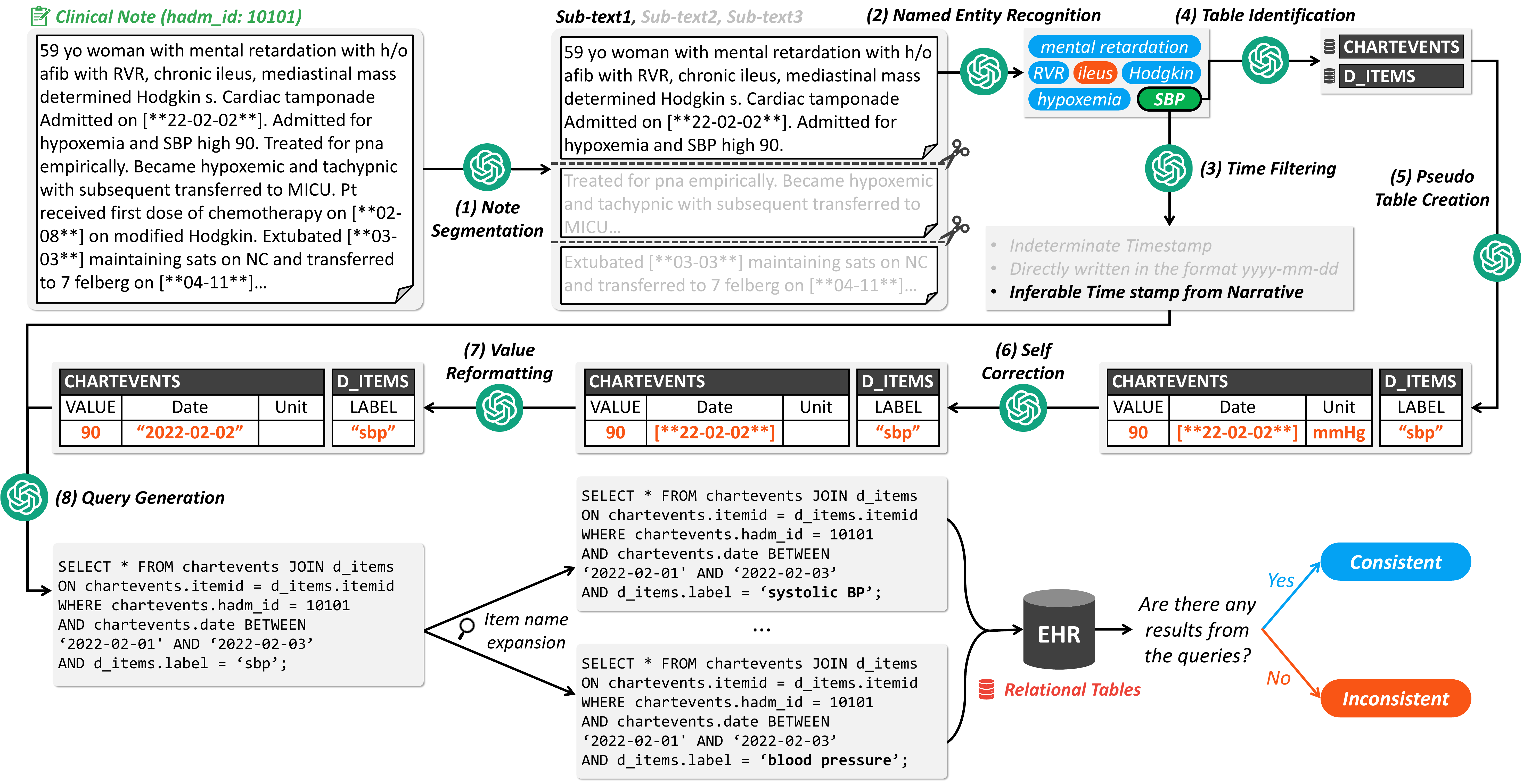}
    \caption{Overview of \mname. The framework consists of eight distinct stages: Note Segmentation, Named Entity Recognition, Time Filtering, Table Identification, Pseudo Table Creation, Self-Correction, Value Reformatting, and Query Generation.}
    \vspace{-2mm} 
    \label{figure3}
    \vspace{-2mm}
\end{figure}

\mname is a novel framework designed to automatically verify the consistency between clinical notes and a relational database. 
As depicted in Figure~\ref{figure3}, \mname encompasses eight sequential stages: Note Segmentation, Named Entity Recognition (NER), Time Filtering, Table Identification, Pseudo Table Creation, Self Correction, Value Reformatting, and Query Generation. 
All stages utilize the in-context learning method with a few examples to maximize the reasoning ability of large language models (LLMs).
The prompts for each step are included in Appendix \ref{app: prompt}.

\textbf{Note Segmentation}~~
LLMs face significant challenges in processing long clinical notes due to its limitations in handling extensive context lengths. 
To overcome this challenge, we propose a new scalable method called Note Segmentation, which divides the entire clinical note into smaller sub-texts that each focus on a specific topic. The following outlines the process of creating a set $\mathcal{T}$, composed of sub-texts from clinical note $P$.
First, the text $P$ is divided into two parts: $P_0^f$, containing the first $l$ tokens, and the remaining text, $P_0^b$. Then, $P_0^f$ is segmented by the LLM into $n$ sub-texts, each with its own distinct topic: \{$P_{0,1}^f$, $P_{0,2}^f$, ..., $P_{0,n}^f$\}.\footnote{$l$ is determined by the context length of the LLM. In this study, $l$ was set to 1000 and $n$ to 3.}
The sub-texts from $P_{0,1}^f$ to $P_{0,n-1}^f$ are added to the set $\mathcal{T}$.
Since $P_{0,n}^f$ is likely incomplete due to the $l$ token limit, it is concatenated with $P_0^b$ for further segmentation.
The combined text of $P_{0,n}^f$ and $P_0^b$ is referred to as $P_1$. This segmentation continues until the length of $P_i$ is $l$ tokens or less, at which point $P_{i}$ is added to $\mathcal{T}$. To ensure smooth transitions, each sub-text includes some content from adjacent sub-texts.
The algorithm and conceptual figure of Note Segmentation are detailed in Appendix~\ref{app: algorithm}.

\textbf{Named Entity Recognition}~~
Our task takes the entire text as input, making the extraction of named entities essential for consistency checks. 
In this task, the LLM extracts entities related to the 13 tables, focusing on those with clear numeric values, and those whose existence can be verified in the database even without explicit values.\footnote{
Narrowing down the NER target like this might seem like taking advantage of our knowledge from the dataset construction process. 
However, we would like to emphasize that the scope of named entities is part of the task definition, and it is essential to share this information with the model so that the model at least understands the objective.
}
This selective extraction is crucial for maintaining the accuracy and reliability of our checks.

\textbf{Time Filtering}~~
At this stage, the LLM determines whether the time expression of a clinical event is in a specific time format, written in a narrative style, or if the time is not specified. The results from this step are utilized for generating queries at the last stage.

\textbf{Table Identification}~~To create a pseudo table in the next stage, it is essential to identify the relevant tables related to the entities. At this stage, the LLM uses table descriptions, foreign key relationships, and just two example rows to identify the necessary table names.

\textbf{Pseudo Table Creation}~~Since clinical notes include content that cannot be easily verified through tables, the LLM creates a pseudo table to effectively extract table-related information. The LLM extracts the information through a multi-step process as follows:
First, extracts sentences from the clinical note that contain the entity to verify.
Then, analyzes the extracted sentences to determine the time information of the entity.
Finally, completes the pseudo table by extracting information about the remaining columns (\textit{e.g}., value, unit) from the notes, using the previously obtained information.
Examples of the detailed process for creating the pseudo table can be found in Appendix \ref{app: pseudo table}.

\textbf{Self Correction}~~During the construction of the pseudo table, we found that hallucinations by the LLM were frequent (see Appendix \ref{app: hallucination}). For example, there were instances where the LLM generated unit information that was not present in the notes.
To address this issue, the LLM re-evaluates whether the pseudo table created in the previous stage is directly aligned with the notes. We then use only the results that are actually aligned.

\textbf{Value Reformatting}~~In clinical notes and tables, the same information may be expressed differently. For instance, a clinical note might mention `\textit{admission}', while the table might record the admission date as `\textit{2022-02-02}'. To align the data types between the generated pseudo table and the actual table, the LLM reformats the pseudo table by using the schema information.

\textbf{Query Generation}~~Using the results from the Time Filtering and Value Reformatting, the LLM creates an SQL query. This query is executed against the database to check if the content mentioned in the notes matches the actual database content.
During execution, we replace the entity in the SQL query with items retrieved from the database. This process involves leveraging the Item Search Tool (see Sec.~\ref{sec3.2}) to cover both medication brand names and their corresponding abbreviations.

\section{Experiments}
\subsection{Experimental Setting}
\textbf{Base LLMs}~~We aim to conduct an evaluation of our \dname and our proposed framework \mname.
For this evaluation, we utilized Tulu2 70B \citep{ivison2023camels}, Mixtral 8X7B \citep{jiang2024mixtral}, Llama-3 70B\footnote{https://llama.meta.com/llama3}, and GPT-3.5 (0613)~\cite{chatgpt} as the base LLMs within our framework.
To effectively measure \mname's performance, experiments were conducted under both few-shot and zero-shot settings. 

\textbf{Note Pre-processing}~~We filtered out information from the notes that is difficult to confirm from the tables (\textit{e.g.}, pre-admission history) to focus on the current admission records (see Appendix \ref{app:notep}). 
All experiments used these processed notes, with results using the unfiltered original notes available in Appendix \ref{app: origianl_note_exp}.

\textbf{Metrics}~~
We evaluate \mname's performance for each note using Precision, Recall and Intersection, then calculate the average across all notes. 
Precision is the number of correctly classified entities divided by the number of all recognized entities\footnote{Note that \mname must first \textit{recognize} entities in the notes during the NER stage before classifying them as \textsc{Consistent} or \textsc{Inconsistent}.}
in the note.
Recall is the number of correctly classified entities divided by the number of all human-labeled entities in the note.
Intersection is the number of correctly classified entities divided by the number of \textit{correctly} recognized entities in the note.
Note that we use Intersection to assess how well \mname performs at least for the correctly recognized entities, considering the difficulty of NER.
Details on the experiment setups are in Appendix \ref{app:evaluate metrics example}.

\renewcommand{\arraystretch}{1.4}
\begin{table}[]
\caption{The main results of \mname on MIMIC-III. Mixtral scored zero in the zero-shot setting, so it was not included in the table. Values in \textbf{bold} represent the highest performance for each metric among all models within the same shot setting.}
\resizebox{\columnwidth}{!}{%
\begin{tabular}{ccccclccclccclccc}
\Xhline{0.3mm}
\multirow{2}{*}{\textbf{Shot}} & \multirow{2}{*}{\textbf{Models}} & \multicolumn{3}{c}{\textbf{Discharge Summary}} & \textbf{} & \multicolumn{3}{c}{\textbf{Physician Note}} & \textbf{} & \multicolumn{3}{c}{\textbf{Nursing Note}} & \textbf{} & \multicolumn{3}{c}{\textbf{Total}} \\ \cline{3-5} \cline{7-9} \cline{11-13} \cline{15-17} 
 &  & Rec & Prec & \multicolumn{1}{l}{Inters} &  & Rec & Prec & \multicolumn{1}{l}{Inters} &  & Rec & Prec & \multicolumn{1}{l}{Inters} &  & Rec & Prec & \multicolumn{1}{l}{Inters} \\ \hline
\multirow{4}{*}{Zero} & Tulu2 & 11.82 & 27.48 & 46.92 &  & 9.1 & 20.15 & 40.83 &  & 15.32 & 23.23 & 30.37 &  & 12.08 & 23.62 & 38.37 \\
 & Mixtral & - & - & - &  & - & - & - &  & - & - & - &  & - & - & - \\
 & Llama-3 & 50.82 & 35.54 & 69.70 &  & 52.92 & 33.89 & 72.71 &  & 53.45 & 44.61 & 81.48 &  & \textbf{52.39} & 38.01 & \textbf{74.03} \\
 &  GPT-3.5 (0613) & 45.04 & 46.71 & 74.58 &  & 40.14 & 37.32 & 70.07 &  & 43.30 & 44.53 & 70.81 &  & 42.83 & \textbf{42.85} & 71.82 \\ \hline
\multirow{4}{*}{Few} & Tulu2 & 40.01 & 49.42 & 70.66 &  & 49.98 & 47.08 & 85.33 &  & 44.77& 40.50 & 78.40 &  & 44.95 & 45.66 & 78.13 \\
 & Mixtral & 54.70 & 49.76 & 71.21 &  & 53.71 & 37.97 & 83.48 &  & 69.86 & 49.65 & 85.01 &  & 54.70 & 45.79 & 79.90 \\
 & Llama-3 & 50.44 & 47.01 & 76.25 &  & 56.11 & 42.75 & 84.30 &  & 52.60 & 38.08 & 75.96 &  & 53.05 & 42.61 & 78.83 \\
 &  GPT-3.5 (0613) & 64.31 & 54.64 & 81.60 &  & 54.64 & 44.01 & 81.41 &  &  64.25 & 47.25 & 95.74 &  &\textbf{61.06} & \textbf{48.63} & \textbf{86.25} \\ 
 \Xhline{0.3mm}
\end{tabular}%
}\label{main}
\vspace{-3mm}
\end{table}
\subsection{Results}
Table \ref{main} presents the results of our framework for both the few-shot and zero-shot settings.
Notably, using GPT-3.5 (0613) as the base LLM in the few-shot scenario achieves the best result, with a recall of 61.06\%, precision of 48.63\%, and an intersection score of 86.25\%. 
This result demonstrates significantly improved performance compared to the direct use of GPT-3.5\footnote{We provided GPT-3.5 (0613) with all the necessary information (\textit{e.g.,} column name, table name, column descriptions) for verification and conducted the consistency check directly.}, which had a recall of 10.12\% and a precision of 8.75\%.

However, despite the carefully crafted 8-stage \mname framework, the overall recall scores remain in the 40-60\% range, underscoring the inherent difficulty of the task.
This challenge is further underscored by the significant gaps between recall and intersection, and between precision and intersection.
Such discrepancies indicate the difficulty of NER in our task. 
Despite providing the model all the entity extraction criteria defined for the task during the NER stage, both recall and precision performance remains low.
This suggests that LLMs lack capabilities required to comprehend clinical notes and accurately extract only the entities that meet the criteria.

Furthermore, in our comparative analysis of zero-shot and few-shot performance, we observed significant improvements with few-shot examples in models like Tulu2, Mixtral, and GPT-3.5 (0613).
However, Llama-3 exhibits similar performance in both zero-shot and few-shot settings. 
Interestingly, few-shot samples improves Llama-3's performance for discharge summaries and physician notes, but degrades for nursing notes.
This suggests that Llama-3 struggles to derive general patterns from in-context examples, particularly in more unstructured formats. Discharge summaries and physician notes typically contain semi-structured patterns (\textit{e.g.}, ``\textit{[2022-02-02 04:06:00] WBC - 9.6}''), making it easier for models to generalize from in-context examples. In contrast, nursing notes are often written in free-form text, presenting a challenge for Llama-3 to generalize effectively from few-shot samples.

\begin{table}[]
\centering
\caption{The main results of CheckEHR on MIMIC-OMOP. In this experiment, the NER step was skipped, and the gold entity was provided. The experiment was conducted in a few-shot setting. Values in \textbf{bold} represent the Total Recall and Total Precision from the model that performed better across the MIMIC-OMOP and MIMIC-III datasets.}
\resizebox{0.9\columnwidth}{!}{%
\begin{tabular}{cccclcclcclcc}
\Xhline{0.3mm}
\multirow{2}{*}{\textbf{Data}} & \multirow{2}{*}{\textbf{Models}} & \multicolumn{2}{c}{\textbf{Discharge}} & \textbf{} & \multicolumn{2}{c}{\textbf{Physician}} & \textbf{} & \multicolumn{2}{c}{\textbf{Nursing}} & \textbf{} & \multicolumn{2}{c}{\textbf{Total}} \\ \cline{3-4} \cline{6-7} \cline{9-10} \cline{12-13} 
 &  & \textbf{Rec} & \textbf{Prec} & \textbf{} & \textbf{Rec} & \textbf{Prec} & \textbf{} & \textbf{Rec} & \textbf{Prec} & \textbf{} & \textbf{Rec} & \textbf{Prec} \\ \hline
\multirow{3}{*}{\textbf{MIMIC-OMOP}} & Tulu2 & 56.08 & 55.23 &  & 53.78 & 61.79 &  & 52.23 & 52.28 &  & \textbf{54.36} & 56.43 \\
 & Mixtral & 53.20 & 54.86 &  & 48.50 & 63.90 &  & 58.16 & 59.29 &  & 53.28 & 59.35 \\
 & Llama-3 & 53.83 & 58.22 &  & 53.82 & 76.95 &  & 59.49 & 60.06 &  & 55.71 & \textbf{65.07} \\ \hline
\multirow{3}{*}{\textbf{MIMIC-III}} & Tulu2 & 55.72 & 66.04 &  & 55.44 & 68.42 &  & 53.23 & 68.18 &  & 54.13 & \textbf{67.54} \\
 & Mixtral & 55.23 & 65.20 &  & 63.55 & 57.78 &  & 69.62 & 68.25 &  & \textbf{62.80} & \textbf{63.74} \\
 & Llama-3 & 53.85 & 63.28 &  & 68.19 & 64.45 &  & 64.25 & 63.35 &  & \textbf{62.09} & 63.69 \\ \Xhline{0.3mm}
\end{tabular}%
}\label{omop_result}
\vspace{-5mm}
\end{table}
\subsection{Result in MIMIC-OMOP}
The MIMIC-III database stores various clinical events in multiple tables like \textit{``chartevents''} and \textit{``labevents''}, while the OMOP CDM organizes these events into standardized tables such as the \textit{``measurements''} table, facilitating multi-organization biomedical research.
Additionally, MIMIC-III uses a variety of dictionary tables (\textit{e.g.}, d\_items, d\_icd\_diagnoses), while the OMOP CDM uses only the \textit{``concept''} table for this purpose (see Appendix \ref{machig_mimic_omop}).
This characteristic of the OMOP CDM simplifies table identification and item search within the database, leading us to anticipate superior performance of MIMIC-OMOP over MIMIC-III.
Contrary to our expectations, however, the performance on MIMIC-OMOP was found to be similar to or lower than that of MIMIC, as shown in Table \ref{omop_result}. 
To understand this discrepancy, we conducted an analysis based on entity types (see Sec.~\ref{sec3.1}) and discovered that a significant performance drop occurred with Type 1 entities.
This decline was mainly caused by the complexity of entities within the MIMIC-OMOP database.
MIMIC-OMOP includes detailed and diverse information related to specific entity names, such as \textit{``Cipralex 10mg tablets (Sigma Pharmaceuticals Plc) 28 tablets''}, encompassing value, unit, and other related details all at once.
Our findings indicate the necessity for developing a framework that can freely interact with the database to overcome these challenges in future research.
The detailed experimental results for each type of OMOP and MIMIC-III can be found in Appendix \ref{app:type exp}.
\subsection{Component Analysis}
For evaluating the role of each component of our framework, we performed further analysis on 25\% of the entire test set. This analysis involved three distinct experimental settings. In the first experimental setting, we excluded the NER stage (Figure~\ref{figure3}-2) and provided the ground truth entities. The second setting built upon this by adding ground truth for the time filtering and table identification (Figure~\ref{figure3}-3,4) stages. In the third setting, we further included ground truth for pseudo table creation, self correction, and value reformatting stage (Figure~\ref{figure3}-5,6,7). 
According to Table \ref{app: model_other1}, our experiments with GPT-3.5 (0613) demonstrated a significant improvement in recall:  76.11\% at the first setting, 82.49\% at the second, and 92.83\% at the third, with an approximate increase of 8 percentage points at each setting.
This finding indicates that the information provided at each stage plays a crucial role in enabling the model to better understand and solve the task. Notably, the performance in the third setting exceeded 92\%, showing a significant improvement over the second setting, indicating that LLMs struggle considerably with converting free text into a structured format.
Refer to Appendix \ref{app: component analysis} for experimental results and additional analysis.

\section{Conclusion and Future Direction}
In this paper, we introduce \dname, a carefully crafted dataset designed to improve the accuracy and reliability of EHRs. By meticulously comparing clinical notes with their corresponding database, \dname addresses critical inconsistencies that can jeopardize patient safety and care quality. Alongside \dname, we present \mname, an innovative framework that leverages LLMs to efficiently verify data consistency within EHRs. Our study lays the groundwork for future advancements in automated and dependable healthcare documentation systems, ultimately enhancing patient safety and streamlining healthcare processes.

Despite the careful design of our dataset, several limitations exist. First, although MIMIC-III is hospital data, preprocessing is required to protect patient privacy. This preprocessing can introduce inconsistencies that do not occur in the actual hospital setting. Therefore, the inconsistencies we identified may not be present in real hospital data. In this regard, future research should incorporate consistency checks using real hospital data to identify inconsistency patterns in practical settings. Secondly, despite the high quality of our dataset, created by highly trained human annotators, there are limitations in verifying the contents of all clinical notes in MIMIC-III. To cover a broader range of cases, more scalable methods will be required.

\label{conclu}

\begin{ack}
This work was supported by the Institute of Information \& Communications Technology Planning \& Evaluation (IITP) grant (No.RS-2019-II190075), National Research Foundation of Korea (NRF) grant (NRF-2020H1D3A2A03100945), and the Korea Health Industry Development Institute (KHIDI) grant (No.HR21C0198, No.HI22C0452), funded by the Korea government (MSIT, MOHW).
\end{ack}

\medskip

{
\small

\bibliographystyle{plain}
\bibliography{custom}

\begin{thebibliography}{10}

\bibitem{aly2021feverous}
Rami Aly, Zhijiang Guo, Michael Schlichtkrull, James Thorne, Andreas Vlachos, Christos Christodoulopoulos, Oana Cocarascu, and Arpit Mittal.
\newblock Feverous: Fact extraction and verification over unstructured and structured information.
\newblock {\em arXiv preprint arXiv:2106.05707}, 2021.

\bibitem{anil2023palm}
Rohan Anil, Andrew~M Dai, Orhan Firat, Melvin Johnson, Dmitry Lepikhin, Alexandre Passos, Siamak Shakeri, Emanuel Taropa, Paige Bailey, Zhifeng Chen, et~al.
\newblock Palm 2 technical report.
\newblock {\em arXiv preprint arXiv:2305.10403}, 2023.

\bibitem{aziz2015electronic}
Hassan~A Aziz and Ola~Asaad Alsharabasi.
\newblock Electronic health records uses and malpractice risks.
\newblock {\em Clinical Laboratory Science}, 28(4):250--255, 2015.

\bibitem{bell2020frequency}
Sigall~K Bell, Tom Delbanco, Joann~G Elmore, Patricia~S Fitzgerald, Alan Fossa, Kendall Harcourt, Suzanne~G Leveille, Thomas~H Payne, Rebecca~A Stametz, Jan Walker, et~al.
\newblock Frequency and types of patient-reported errors in electronic health record ambulatory care notes.
\newblock {\em JAMA network open}, 3(6):e205867--e205867, 2020.

\bibitem{black2017transitions}
Meghan Black and Cristin~M Colford.
\newblock Transitions of care: improving the quality of discharge summaries completed by internal medicine residents.
\newblock {\em MedEdPORTAL}, 13:10613, 2017.

\bibitem{brown2020language}
Tom Brown, Benjamin Mann, Nick Ryder, Melanie Subbiah, Jared~D Kaplan, Prafulla Dhariwal, Arvind Neelakantan, Pranav Shyam, Girish Sastry, Amanda Askell, et~al.
\newblock Language models are few-shot learners.
\newblock {\em Advances in neural information processing systems}, 33:1877--1901, 2020.

\bibitem{chen2019tabfact}
Wenhu Chen, Hongmin Wang, Jianshu Chen, Yunkai Zhang, Hong Wang, Shiyang Li, Xiyou Zhou, and William~Yang Wang.
\newblock Tabfact: A large-scale dataset for table-based fact verification.
\newblock {\em arXiv preprint arXiv:1909.02164}, 2019.

\bibitem{gupta-etal-2020-infotabs}
Vivek Gupta, Maitrey Mehta, Pegah Nokhiz, and Vivek Srikumar.
\newblock {INFOTABS}: Inference on tables as semi-structured data.
\newblock In Dan Jurafsky, Joyce Chai, Natalie Schluter, and Joel Tetreault, editors, {\em Proceedings of the 58th Annual Meeting of the Association for Computational Linguistics}, pages 2309--2324, Online, July 2020. Association for Computational Linguistics.

\bibitem{hammerling2012review}
Julie~A Hammerling.
\newblock A review of medical errors in laboratory diagnostics and where we are today.
\newblock {\em Laboratory medicine}, 43(2):41--44, 2012.

\bibitem{ivison2023camels}
Hamish Ivison, Yizhong Wang, Valentina Pyatkin, Nathan Lambert, Matthew Peters, Pradeep Dasigi, Joel Jang, David Wadden, Noah~A Smith, Iz~Beltagy, et~al.
\newblock Camels in a changing climate: Enhancing lm adaptation with tulu 2.
\newblock {\em arXiv preprint arXiv:2311.10702}, 2023.

\bibitem{jiang2024mixtral}
Albert~Q Jiang, Alexandre Sablayrolles, Antoine Roux, Arthur Mensch, Blanche Savary, Chris Bamford, Devendra~Singh Chaplot, Diego de~las Casas, Emma~Bou Hanna, Florian Bressand, et~al.
\newblock Mixtral of experts.
\newblock {\em arXiv preprint arXiv:2401.04088}, 2024.

\bibitem{jiang-etal-2023-structgpt}
Jinhao Jiang, Kun Zhou, Zican Dong, Keming Ye, Xin Zhao, and Ji-Rong Wen.
\newblock {S}truct{GPT}: A general framework for large language model to reason over structured data.
\newblock In {\em Proceedings of the 2023 Conference on Empirical Methods in Natural Language Processing}, pages 9237--9251, Singapore, December 2023. Association for Computational Linguistics.

\bibitem{jiang2020hover}
Yichen Jiang, Shikha Bordia, Zheng Zhong, Charles Dognin, Maneesh Singh, and Mohit Bansal.
\newblock Hover: A dataset for many-hop fact extraction and claim verification.
\newblock {\em arXiv preprint arXiv:2011.03088}, 2020.

\bibitem{johnson2020mimic}
Alistair Johnson, Lucas Bulgarelli, Tom Pollard, Steven Horng, Leo~Anthony Celi, and Roger Mark.
\newblock Mimic-iv.
\newblock {\em PhysioNet. Available online at: https://physionet. org/content/mimiciv/1.0/(accessed August 23, 2021)}, pages 49--55, 2020.

\bibitem{johnson2016physionet}
Alistair E.~W. Johnson, Tom~J. Pollard, and Roger~G. Mark.
\newblock {MIMIC-III} clinical database (version 1.4), 2016.

\bibitem{khot2022decomposed}
Tushar Khot, Harsh Trivedi, Matthew Finlayson, Yao Fu, Kyle Richardson, Peter Clark, and Ashish Sabharwal.
\newblock Decomposed prompting: A modular approach for solving complex tasks.
\newblock In {\em The Eleventh International Conference on Learning Representations}, 2022.

\bibitem{kim-etal-2023-kg}
Jiho Kim, Yeonsu Kwon, Yohan Jo, and Edward Choi.
\newblock {KG}-{GPT}: A general framework for reasoning on knowledge graphs using large language models.
\newblock In {\em Findings of the Association for Computational Linguistics: EMNLP 2023}, pages 9410--9421, Singapore, December 2023. Association for Computational Linguistics.

\bibitem{kim-etal-2023-factkg}
Jiho Kim, Sungjin Park, Yeonsu Kwon, Yohan Jo, James Thorne, and Edward Choi.
\newblock {F}act{KG}: Fact verification via reasoning on knowledge graphs.
\newblock In {\em Proceedings of the 61st Annual Meeting of the Association for Computational Linguistics (Volume 1: Long Papers)}, pages 16190--16206, Toronto, Canada, July 2023. Association for Computational Linguistics.

\bibitem{lu2024chameleon}
Pan Lu, Baolin Peng, Hao Cheng, Michel Galley, Kai-Wei Chang, Ying~Nian Wu, Song-Chun Zhu, and Jianfeng Gao.
\newblock Chameleon: Plug-and-play compositional reasoning with large language models.
\newblock {\em Advances in Neural Information Processing Systems}, 36, 2024.

\bibitem{openai2023gpt4}
OpenAI.
\newblock Gpt-4 technical report, 2023.

\bibitem{chatgpt}
OpenAI.
\newblock Introducing chatgpt., 2023.

\bibitem{palchuk2010unintended}
Matvey~B Palchuk, Elizabeth~A Fang, Janet~M Cygielnik, Matthew Labreche, Maria Shubina, Harley~Z Ramelson, Claus Hamann, Carol Broverman, Jonathan~S Einbinder, and Alexander Turchin.
\newblock An unintended consequence of electronic prescriptions: prevalence and impact of internal discrepancies.
\newblock {\em Journal of the American Medical Informatics Association}, 17(4):472--476, 2010.

\bibitem{park2021faviq}
Jungsoo Park, Sewon Min, Jaewoo Kang, Luke Zettlemoyer, and Hannaneh Hajishirzi.
\newblock Faviq: Fact verification from information-seeking questions.
\newblock {\em arXiv preprint arXiv:2107.02153}, 2021.

\bibitem{payne2018using}
Thomas~H Payne, W~David Alonso, J~Andrew Markiel, Kevin Lybarger, Ross Lordon, Meliha Yetisgen, Jennifer~M Zech, and Andrew~A White.
\newblock Using voice to create inpatient progress notes: effects on note timeliness, quality, and physician satisfaction.
\newblock {\em JAMIA open}, 1(2):218--226, 2018.

\bibitem{c4wsrs}
Alvin Rajkomar, Eric Loreaux, Yuchen Liu, Jonas Kemp, Benny Li, Ming-Jun Chen, Yi~Zhang, Afroz Mohiuddin, and Juraj Gottweis.
\newblock Deciphering clinical abbreviations with a privacy protecting machine learning system.
\newblock {\em Nature Communications}, 13, 12 2022.

\bibitem{rani-etal-2023-factify}
Anku Rani, S.M Towhidul~Islam Tonmoy, Dwip Dalal, Shreya Gautam, Megha Chakraborty, Aman Chadha, Amit Sheth, and Amitava Das.
\newblock {FACTIFY}-5{WQA}: 5{W} aspect-based fact verification through question answering.
\newblock In Anna Rogers, Jordan Boyd-Graber, and Naoaki Okazaki, editors, {\em Proceedings of the 61st Annual Meeting of the Association for Computational Linguistics (Volume 1: Long Papers)}, pages 10421--10440, Toronto, Canada, July 2023. Association for Computational Linguistics.

\bibitem{saakyan2021covid}
Arkadiy Saakyan, Tuhin Chakrabarty, and Smaranda Muresan.
\newblock Covid-fact: Fact extraction and verification of real-world claims on covid-19 pandemic.
\newblock {\em arXiv preprint arXiv:2106.03794}, 2021.

\bibitem{sarrouti2021evidence}
Mourad Sarrouti, Asma~Ben Abacha, Yassine M’rabet, and Dina Demner-Fushman.
\newblock Evidence-based fact-checking of health-related claims.
\newblock In {\em Findings of the Association for Computational Linguistics: EMNLP 2021}, pages 3499--3512, 2021.

\bibitem{schuster2021get}
Tal Schuster, Adam Fisch, and Regina Barzilay.
\newblock Get your vitamin c! robust fact verification with contrastive evidence.
\newblock {\em arXiv preprint arXiv:2103.08541}, 2021.

\bibitem{schuster2019towards}
Tal Schuster, Darsh~J Shah, Yun Jie~Serene Yeo, Daniel Filizzola, Enrico Santus, and Regina Barzilay.
\newblock Towards debiasing fact verification models.
\newblock {\em arXiv preprint arXiv:1908.05267}, 2019.

\bibitem{thorne2018fever}
James Thorne, Andreas Vlachos, Christos Christodoulopoulos, and Arpit Mittal.
\newblock Fever: a large-scale dataset for fact extraction and verification.
\newblock {\em arXiv preprint arXiv:1803.05355}, 2018.

\bibitem{villa2014review}
Luis~Bernardo Villa and Ivan Cabezas.
\newblock A review on usability features for designing electronic health records.
\newblock In {\em 2014 IEEE 16th international conference on e-health networking, applications and services (Healthcom)}, pages 49--54. IEEE, 2014.

\bibitem{vladika2023scientific}
Juraj Vladika and Florian Matthes.
\newblock Scientific fact-checking: A survey of resources and approaches.
\newblock {\em arXiv preprint arXiv:2305.16859}, 2023.

\bibitem{omop}
Erica Voss, Rupa Makadia, Amy Matcho, Qianli Ma, Chris Knoll, Martijn Schuemie, Frank Defalco, Ajit Londhe, Vivienne Zhu, and Patrick Ryan.
\newblock Feasibility and utility of applications of the common data model to multiple, disparate observational health databases.
\newblock {\em Journal of the American Medical Informatics Association : JAMIA}, 22, 02 2015.

\bibitem{wang-etal-2021-semeval}
Nancy X.~R. Wang, Diwakar Mahajan, Marina Danilevsky, and Sara Rosenthal.
\newblock {S}em{E}val-2021 task 9: Fact verification and evidence finding for tabular data in scientific documents ({SEM}-{TAB}-{FACTS}).
\newblock In {\em Proceedings of the 15th International Workshop on Semantic Evaluation (SemEval-2021)}, pages 317--326, Online, August 2021.

\bibitem{wang2023chain}
Zilong Wang, Hao Zhang, Chun-Liang Li, Julian~Martin Eisenschlos, Vincent Perot, Zifeng Wang, Lesly Miculicich, Yasuhisa Fujii, Jingbo Shang, Chen-Yu Lee, et~al.
\newblock Chain-of-table: Evolving tables in the reasoning chain for table understanding.
\newblock In {\em The Twelfth International Conference on Learning Representations}, 2023.

\bibitem{ward2015effects}
Michael~J Ward, Wesley~H Self, and Craig~M Froehle.
\newblock Effects of common data errors in electronic health records on emergency department operational performance metrics: A monte carlo simulation.
\newblock {\em Academic Emergency Medicine}, 22(9):1085--1092, 2015.

\bibitem{yadav2017comparison}
Siddhartha Yadav, Noora Kazanji, Narayan KC, Sudarshan Paudel, John Falatko, Sandor Shoichet, Michael Maddens, and Michael~A Barnes.
\newblock Comparison of accuracy of physical examination findings in initial progress notes between paper charts and a newly implemented electronic health record.
\newblock {\em Journal of the American Medical Informatics Association}, 24(1):140--144, 2017.

\bibitem{zhou2022least}
Denny Zhou, Nathanael Sch{\"a}rli, Le~Hou, Jason Wei, Nathan Scales, Xuezhi Wang, Dale Schuurmans, Claire Cui, Olivier Bousquet, Quoc~V Le, et~al.
\newblock Least-to-most prompting enables complex reasoning in large language models.
\newblock In {\em The Eleventh International Conference on Learning Representations}, 2022.

\end{thebibliography}

}

\newpage
\section*{Checklist}

\begin{enumerate}

\item For all authors...
\begin{enumerate}
  \item Do the main claims made in the abstract and introduction accurately reflect the paper's contributions and scope?
    \answerYes{See Section~\ref{intro}.}
  \item Did you describe the limitations of your work?
    \answerYes{See Section~\ref{conclu}.}
  \item Did you discuss any potential negative societal impacts of your work?
    \answerNo{
Our dataset is entirely based on MIMIC-III. All datasets have been de-identified and are provided under the PhysioNet license. Moreover, we have not attempted to identify any clinical records within the dataset to prevent potential negative social consequences.}
  \item Have you read the ethics review guidelines and ensured that your paper conforms to them?
    \answerYes
\end{enumerate}

\item If you are including theoretical results...
\begin{enumerate}
  \item Did you state the full set of assumptions of all theoretical results?
    \answerNA{}
	\item Did you include complete proofs of all theoretical results?
    \answerNA{}
\end{enumerate}

\item If you ran experiments (e.g. for benchmarks)...
\begin{enumerate}
  \item Did you include the code, data, and instructions needed to reproduce the main experimental results (either in the supplemental material or as a URL)?
    \answerYes{See Appendix \ref{app:evaluate metrics example}}. We will release the experimental code.
  \item Did you specify all the training details (e.g., data splits, hyperparameters, how they were chosen)?
    \answerYes{See Appendix \ref{app:evaluate metrics example}}.
	\item Did you report error bars (e.g., with respect to the random seed after running experiments multiple times)?
    \answerNo {We did not conduct the experiment multiple times.}
	\item Did you include the total amount of compute and the type of resources used (e.g., type of GPUs, internal cluster, or cloud provider)?
    \answerYes{See Appendix \ref{app:evaluate metrics example}}.
\end{enumerate}

\item If you are using existing assets (e.g., code, data, models) or curating/releasing new assets...
\begin{enumerate}
  \item If your work uses existing assets, did you cite the creators?
    \answerYes
  \item Did you mention the license of the assets?
    \answerYes
  \item Did you include any new assets either in the supplemental material or as a URL?
    \answerYes{We will share a Google Drive link containing new assets. After all review periods have ended, the link will be deprecated.}
  \item Did you discuss whether and how consent was obtained from people whose data you're using/curating?
    \answerNo
  \item Did you discuss whether the data you are using/curating contains personally identifiable information or offensive content?
    \answerNo{Our dataset is sourced from the MIMIC-III database on PhysioNet and includes de-identified patient records. It contains no personally identifiable information or offensive material.}
\end{enumerate}

\item If you used crowdsourcing or conducted research with human subjects...
\begin{enumerate}
  \item Did you include the full text of instructions given to participants and screenshots, if applicable?
    \answerYes {see Appendix \ref{app: prompt}.}
  \item Did you describe any potential participant risks, with links to Institutional Review Board (IRB) approvals, if applicable?
    \answerNo {We used MIMIC-III, which was approved by the Institutional Review Boards of Beth Israel Deaconess Medical Center (Boston, MA) and Massachusetts Institute of Technology (Cambridge, MA).}
  \item Did you include the estimated hourly wage paid to participants and the total amount spent on participant compensation?
    \answerNA
\end{enumerate}

\end{enumerate}


\appendix

\newpage
\startcontents[supplementary]
\printcontents[supplementary]{l}{1}{\section*{Supplementary Contents}}

\newpage
\section{Datasheet for Datasets}
\subsection{Motivation}
\begin{itemize}[label=\textbullet, leftmargin=1cm, labelsep=0.3cm]
\item \textbf{For what purpose was the dataset created?}

EHRs are integral for storing comprehensive patient medical records, combining structured data with detailed clinical notes. However, they often suffer from discrepancies due to unintuitive EHR system designs and human errors, posing serious risks to patient safety. To address this, we developed \dname.

\item \textbf{Who created the dataset (e.g., which team, research group) and on behalf of which entity (e.g., company, institution, organization)?}

The authors created the dataset.

\item \textbf{Who funded the creation of the dataset? If there is an associated grant, please provide
the name of the grantor and the grant name and number.}

This work was supported by the Institute of Information \& Communications Technology Planning \& Evaluation (IITP) grant (No.RS-2019-II190075), National Research Foundation of Korea (NRF) grant (NRF-2020H1D3A2A03100945), and the Korea Health Industry Development Institute (KHIDI) grant (No.HR21C0198, No.HI22C0452), funded by the Korea government (MSIT, MOHW).

\end{itemize}

\subsection{Composition}
\begin{itemize}[label=\textbullet, leftmargin=1cm, labelsep=0.3cm]
\item \textbf{What do the instances that comprise the dataset represent (e.g., documents, photos, people, countries)?}

\dname includes entities identified in the notes along with their labels. Additionally, for inconsistent entities, it identifies the specific table and column in the EHR where the inconsistencies are found.

\item \textbf{How many instances are there in total (of each type, if appropriate)?}

It includes 4,101 entities extracted from a total of 105 clinical notes.

\item \textbf{Does the dataset contain all possible instances or is it a sample (not necessarily random) of instances from a larger set?}

We randomly extracted and used 105 clinical notes from MIMIC-III. The notes consist of discharge summaries, physician notes, and nursing notes.

\item \textbf{What data does each instance consist of?}

Entities extracted from the notes have corresponding labels, and for inconsistent entities, the specific tables and columns in the EHR where the inconsistency occurs are recorded.

\item \textbf{Is there a label or target associated with each instance?}

Each entity has a corresponding label.

\item \textbf{Is any information missing from individual instances? If so, please provide a description, explaining why this information is missing (e.g., because it was unavailable). This does not include intentionally removed information, but might include, e.g., redacted text.}

No.

\item \textbf{Are relationships between individual instances made explicit (e.g., users’ movie ratings, social network links)?}

No.

\item \textbf{Are there recommended data splits (e.g., training, development/validation, testing)?}

We randomly divided 105 clinical notes into a test set of 83 and a validation set of 22 for the experiment.

\item \textbf{Are there any errors, sources of noise, or redundancies in the dataset?}

Although trained human annotators followed the labeling instructions, slight variations may exist due to individual perspectives.

\item \textbf{Is the dataset self-contained, or does it link to or otherwise rely on external resources (e.g., websites, tweets, other datasets)?}

\dname depends on MIMIC-III which is accessible via PhysioNet\footnote{\url{https://physionet.org/}}.

\item \textbf{Does the dataset contain data that might be considered confidential (e.g., data that is protected by legal privilege or by doctor-patient confidentiality, data that includes the content of individuals’ non-public communications)?}

No.

\item \textbf{Does the dataset contain data that, if viewed directly, might be offensive, insulting, threatening, or might otherwise cause anxiety?}

No.

\item \textbf{Does the dataset relate to people?}

Yes.

\item \textbf{Does the dataset identify any subpopulations (e.g., by age, gender)?}

No.

\item \textbf{Does the dataset contain data that might be considered sensitive in any way (e.g., data that reveals race or ethnic origins, sexual orientations, religious beliefs, political opinions or union memberships, or locations; financial or health data; biometric or genetic data; forms of government identification, such as social security numbers; criminal history)?}

No.

\end{itemize}

\subsection{Collection process}
\begin{itemize}[label=\textbullet, leftmargin=1cm, labelsep=0.3cm]
\item \textbf{How was the data associated with each instance acquired?}

Before starting the labeling process, we designed the labeling instructions in consultation with the partitioners. Based on these instructions, trained human annotators reviewed the clinical notes and labeled the entities.

\item \textbf{What mechanisms or procedures were used to collect the data (e.g., hardware apparatuses or sensors, manual human curation, software programs, software APIs)?}

We used Google Search and ChatGPT4 API (Azure) to analyze the entities in clinical notes, and SQLite3 for SQL query execution.

\item \textbf{If the dataset is a sample from a larger set, what was the sampling strategy (e.g., deterministic, probabilistic with specific sampling probabilities)?}

When extracting notes, we randomly selected notes with at least 800 tokens to ensure they contained sufficient content.

\item \textbf{Who was involved in the data collection process (e.g., students, crowd workers, contractors) and how were they compensated (e.g., how much were crowd workers paid)?}

The authors manually labeled the data.

\item \textbf{Over what timeframe was the data collected?}

We created the dataset from April 2023 to April 2024.

\item \textbf{Were any ethical review processes conducted (e.g., by an institutional review board)?}

N/A.

\item \textbf{Does the dataset relate to people?}

Yes.

\item \textbf{Did you collect the data from the individuals in question directly, or obtain it via third parties or other sources (e.g., websites)?}

N/A.

\item \textbf{Were the individuals in question notified about the data collection?}

N/A.

\item \textbf{Did the individuals in question consent to the collection and use of their data?}

N/A.

\item \textbf{If consent was obtained, were the consenting individuals provided with a mechanism to revoke their consent in the future or for certain uses?}

N/A.

\item \textbf{Has an analysis of the potential impact of the dataset and its use on data subjects (e.g., a data protection impact analysis) been conducted?}

Yes.

\end{itemize}

\subsection{Preprocessing/cleaning/labeling}
\begin{itemize}[label=\textbullet, leftmargin=1cm, labelsep=0.3cm]

\item \textbf{Was any preprocessing/cleaning/labeling of the data done (e.g., discretization or bucketing, tokenization, part-of-speech tagging, SIFT feature extraction, removal of instances, processing of missing values)?}

We performed preprocessing by removing pre-admission records and treatment plans from the clinical notes.

\item \textbf{Was the ``raw'' data saved in addition to the preprocess/cleaned/labeled data (e.g., to support unanticipated future uses)?}

We additionally provide labels for the original notes.

\item \textbf{Is the software that was used to preprocess/clean/label the data available?}

We performed preprocessing using Python.

\end{itemize}

\subsection{Uses}
\begin{itemize}[label=\textbullet, leftmargin=1cm, labelsep=0.3cm]
\item \textbf{Has the dataset been used for any tasks already?}

No.

\item \textbf{Is there a repository that links to any or all papers or systems that use the dataset?}

N/A.

\item \textbf{What (other) tasks could the dataset be used for?}

It can be used not only for consistency checks of EHR but also for table-based fact verification tasks.

\item \textbf{Is there anything about the composition of the dataset or the way it was collected and preprocessed/cleaned/labeled that might impact future uses?}

N/A.

\item \textbf{Are there tasks for which the dataset should not be used?}

N/A.

\end{itemize}

\subsection{Distribution}
\begin{itemize}[label=\textbullet, leftmargin=1cm, labelsep=0.3cm]
\item \textbf{Will the dataset be distributed to third parties outside of the entity (e.g., company, institution, organization) on behalf of which the dataset was created?}

No.

\item \textbf{How will the dataset be distributed?}

The dataset will be released at PhysioNet.

\item \textbf{Will the dataset be distributed under a copyright or other intellectual property (IP) license, and/or under applicable terms of use (ToU)?}

The dataset is released under MIT License.

\item \textbf{Have any third parties imposed IP-based or other restrictions on the data associated with the instances?}

No.

\item \textbf{Do any export controls or other regulatory restrictions apply to the dataset or to individual instances?}

No.

\end{itemize}

\subsection{Maintenance}
\begin{itemize}[label=\textbullet, leftmargin=1cm, labelsep=0.3cm]
\item \textbf{Who will be supporting/hosting/maintaining the dataset?}

The authors will support it.

\item \textbf{How can the owner/curator/manager of the dataset be contacted(e.g., email address)?}

Contact the authors ({\{yeonsu.k, jiho.kim\}@kaist.ac.kr}).

\item \textbf{Is there an erratum?}

No.

\item \textbf{Will the dataset be updated (e.g., to correct labeling erros, add new instances, delete instances)?}

The authors will update the dataset if any corrections are required.

\item \textbf{If the dataset relates to people, are there applicable limits on the retention of the data associated with the instances (e.g., were the individuals in question told that their data would be retained for a fixed period of time and then deleted)?}

N/A

\item \textbf{Will older versions of the dataset continue to be supported/hosted/maintained?}

We plan to upload the latest version of the dataset and will document the updates for each version separately.

\item \textbf{If others want to extend/augment/build on/contribute to the dataset, is there a mechanism for them to do so?}

Contact the authors ({\{yeonsu.k, jiho.kim\}@kaist.ac.kr}).
\end{itemize}

\newpage
\section{MIMIC-III and OMOP CDM}
\dname is built on two types of relational databases: MIMIC-III and its version in OMOP CDM. This structure allows us to incorporate various schema types and enhance generalizability. In this section, we will describe in detail the process of creating both MIMIC-III and MIMIC-OMOP.
\label{app:mimic and omop}

\subsection{Data Preparation}
We created \dname using 105 randomly selected clinical notes and 13 tables. In this section, we will provide a detailed description of how the clinical notes and tables were selected and preprocessed.

\subsubsection{Note Preparation}
\label{app:notep}
\begin{figure}[h]
\includegraphics[width=0.8\linewidth]{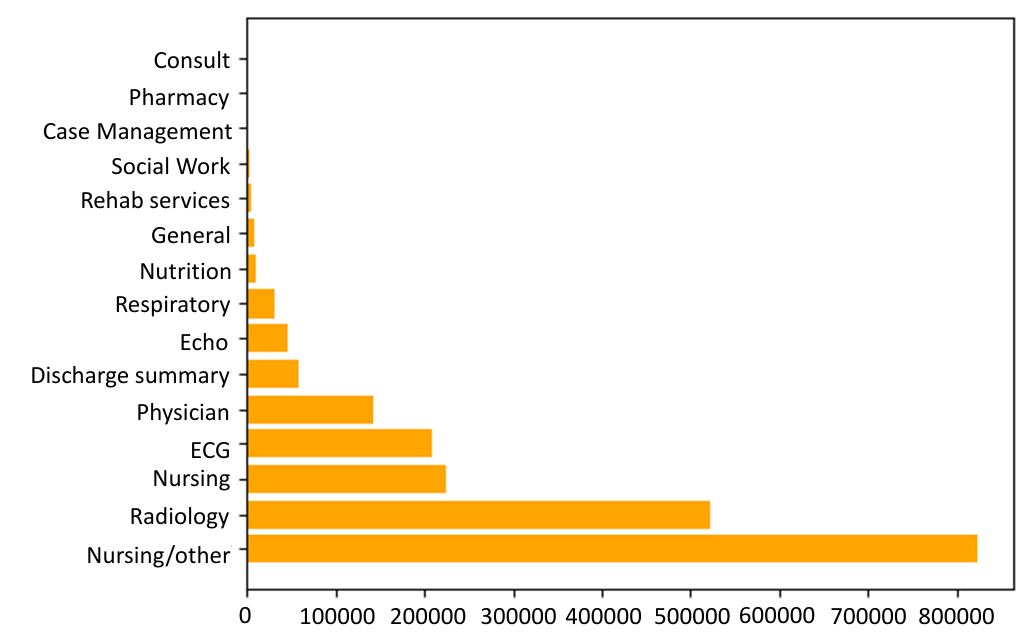}
\vspace{-3mm} 
\caption{Distribution of different note categories in MIMIC-III.}
\label{notestat}
\end{figure}

To develop a more realistic dataset that mirrors a typical hospital setting, we began by analyzing the clinical notes from MIMIC-III by category (see Figure \ref{notestat}). 
The largest portion of notes came from the ``Nursing/other'' category. However, much of this content, such as details of family meetings, couldn't be verified against the table data and was therefore excluded.
Radiology reports and ECG (electrocardiogram) reports were also excluded since they rely on imaging and cardiac monitoring, which are outside our scope. Therefore, our focus was on discharge summaries, nursing notes, and physician notes, as these are commonly used in hospitals and related to tabular information.

Additionally, clinical notes may contain pre-admission history and future treatment plans not found in the EHR tables. To concentrate on current admission records, we filtered out this additional information from the notes. To ensure sufficient detail, we randomly selected 105 notes with more than 800 tokens for the experiments.

\subsubsection{Table Preparation}
\begin{figure}[t]
\includegraphics[width=\linewidth]{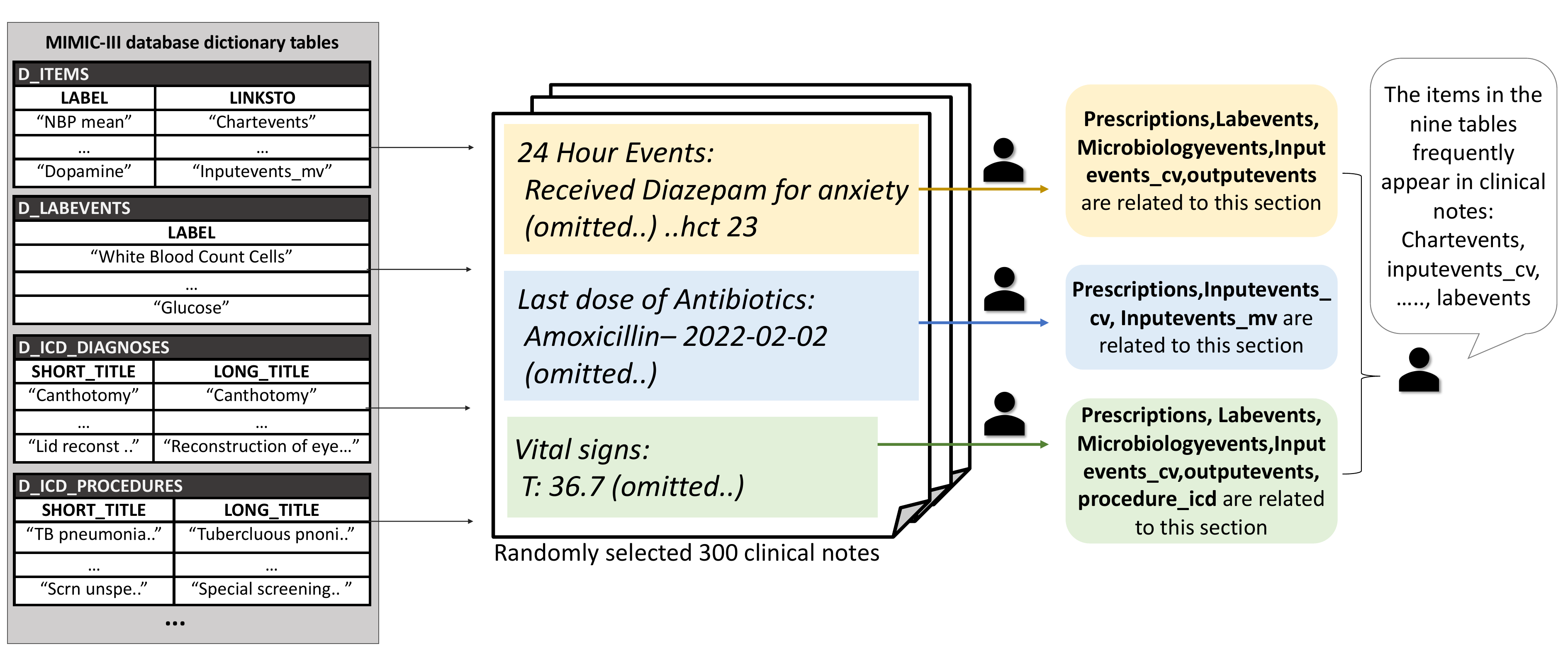}
\caption{Overall process of table preparation.}
    \label{overall_process_table_process}
 \end{figure}

To enhance the utility of our dataset, we identified tables in the MIMIC-III database that contain entities from discharge summaries, physician notes, and nursing notes.
To achieve this, we analyzed randomly selected 300 clinical notes, consisting of 100 discharge summaries, 100 nursing notes, and 100 physician notes. As a result, we concluded with a total of thirteen tables, including four dictionary tables, as follows: Chartevents, Labevents, Prescriptions, Inputevents\_cv, Inputevents\_mv, Outputevents, Microbiologyevents, Diagnoses\_icd, Procedures\_icd, D\_items, D\_icd\_diagnoses, D\_icd\_procedures, and D\_labitems. The overall table preparation process is described in Figure \ref{overall_process_table_process}.

\subsection{OMOP CDM}

\subsubsection{Matching OMOP CDM and MIMIC-III}
Table \ref{table_mapping} illustrates how each MIMIC-III table and column correspond to the respective OMOP CDM table and column.
\label{machig_mimic_omop}
\begin{table}[]
\caption{Relationship between MIMIC-III and MIMIC-OMOP Tables.}
\centering
\renewcommand{\arraystretch}{1.3}
\resizebox{0.75\columnwidth}{!}{%
\begin{tabular}{cc|cc}
\hline
\multicolumn{2}{c|}{\textbf{MIMIC-III}}                                     & \multicolumn{2}{c}{\textbf{MIMIC-OMOP}}                                        \\ \hline
\multicolumn{1}{c|}{\textbf{Table}}                      & \textbf{Column}  & \textbf{Table}                                & \textbf{Column}                \\ \hline
\multicolumn{1}{c|}{\multirow{3}{*}{Chartevents}}        & chartttime                 & \multicolumn{1}{c|}{\multirow{10}{*}{Measurement}}    & measurement\_datetime     \\ \cline{2-2} \cline{4-4} 
\multicolumn{1}{c|}{}                                    & valuenum         & \multicolumn{1}{c|}{}                         & value\_as\_number              \\ \cline{2-2} \cline{4-4} 
\multicolumn{1}{c|}{}                                    & valueuom         & \multicolumn{1}{c|}{}                         & unit\_source\_value            \\ \cline{1-2} \cline{4-4} 
\multicolumn{1}{c|}{\multirow{3}{*}{Labevents}}          & chartttime       & \multicolumn{1}{c|}{}                         & measurement\_datetime          \\ \cline{2-2} \cline{4-4} 
\multicolumn{1}{c|}{}                                    & valuenum         & \multicolumn{1}{c|}{}                         & value\_as\_number              \\ \cline{2-2} \cline{4-4} 
\multicolumn{1}{c|}{}                                    & valueuom         & \multicolumn{1}{c|}{}                         & unit\_source\_value            \\ \cline{1-2} \cline{4-4} 
\multicolumn{1}{c|}{\multirow{3}{*}{Outputevents}}       & chartttime       & \multicolumn{1}{c|}{}                         & measurement\_datetime          \\ \cline{2-2} \cline{4-4} 
\multicolumn{1}{c|}{}                                    & valuenum         & \multicolumn{1}{c|}{}                         & value\_as\_number              \\ \cline{2-2} \cline{4-4} 
\multicolumn{1}{c|}{}                                    & valueuom         & \multicolumn{1}{c|}{}                         & unit\_source\_value            \\ \cline{1-2} \cline{4-4} 
\multicolumn{1}{c|}{\multirow{4}{*}{Microbiologyevents}} & \multirow{2}{*}{charttime} & \multicolumn{1}{c|}{}                                 & measurement\_datetime     \\ \cline{3-4} 
\multicolumn{1}{c|}{}                                    &                  & \multicolumn{1}{c|}{Specimen}                 & specimen\_datetime             \\ \cline{2-4} 
\multicolumn{1}{c|}{}                                    & org\_name        & \multicolumn{1}{c|}{\multirow{2}{*}{Concept}} & \multirow{2}{*}{concept\_name} \\ \cline{2-2}
\multicolumn{1}{c|}{}                                    & spec\_type\_desc & \multicolumn{1}{c|}{}                         &                                \\ \hline
\multicolumn{1}{c|}{\multirow{6}{*}{Inputevents\_mv}}    & starttime                  & \multicolumn{1}{c|}{\multirow{16}{*}{Drug\_exposure}} & drug\_exposure\_startdate \\ \cline{2-2} \cline{4-4} 
\multicolumn{1}{c|}{}                                    & endtime          & \multicolumn{1}{c|}{}                         & drug\_exposure\_enddate        \\ \cline{2-2} \cline{4-4} 
\multicolumn{1}{c|}{}                                    & amount           & \multicolumn{1}{c|}{}                         & -                              \\ \cline{2-2} \cline{4-4} 
\multicolumn{1}{c|}{}                                    & amoutuom         & \multicolumn{1}{c|}{}                         & -                              \\ \cline{2-2} \cline{4-4} 
\multicolumn{1}{c|}{}                                    & rate             & \multicolumn{1}{c|}{}                         & -                              \\ \cline{2-2} \cline{4-4} 
\multicolumn{1}{c|}{}                                    & rateuom          & \multicolumn{1}{c|}{}                         & -                              \\ \cline{1-2} \cline{4-4} 
\multicolumn{1}{c|}{\multirow{6}{*}{Inputevents\_cv}}    & \multirow{2}{*}{charttime} & \multicolumn{1}{c|}{}                                 & drug\_exposure\_startdate \\
\multicolumn{1}{c|}{}                                    &                  & \multicolumn{1}{c|}{}                         & drug\_exposure\_enddate        \\ \cline{2-2} \cline{4-4} 
\multicolumn{1}{c|}{}                                    & amount           & \multicolumn{1}{c|}{}                         & -                              \\ \cline{2-2} \cline{4-4} 
\multicolumn{1}{c|}{}                                    & amoutuom         & \multicolumn{1}{c|}{}                         & -                              \\ \cline{2-2} \cline{4-4} 
\multicolumn{1}{c|}{}                                    & rate             & \multicolumn{1}{c|}{}                         & -                              \\ \cline{2-2} \cline{4-4} 
\multicolumn{1}{c|}{}                                    & rateuom          & \multicolumn{1}{c|}{}                         & -                              \\ \cline{1-2} \cline{4-4} 
\multicolumn{1}{c|}{\multirow{5}{*}{Prescriptions}}      & startdate        & \multicolumn{1}{c|}{}                         & drug\_exposure\_startdate      \\ \cline{2-2} \cline{4-4} 
\multicolumn{1}{c|}{}                                    & enddate          & \multicolumn{1}{c|}{}                         & drug\_exposure\_enddate        \\ \cline{2-2} \cline{4-4} 
\multicolumn{1}{c|}{}                                    & dose\_val\_rx    & \multicolumn{1}{c|}{}                         & -                              \\ \cline{2-2} \cline{4-4} 
\multicolumn{1}{c|}{}                                    & dose\_unit\_rx   & \multicolumn{1}{c|}{}                         & -                              \\ \cline{2-4} 
\multicolumn{1}{c|}{}                                    & drug             & \multicolumn{1}{c|}{\multirow{7}{*}{Concept}} & \multirow{7}{*}{concept\_name} \\ \cline{1-2}
\multicolumn{1}{c|}{D\_items}                            & label            & \multicolumn{1}{c|}{}                         &                                \\ \cline{1-2}
\multicolumn{1}{c|}{D\_labitems}                         & label            & \multicolumn{1}{c|}{}                         &                                \\ \cline{1-2}
\multicolumn{1}{c|}{\multirow{2}{*}{D\_icd\_diagnosis}}  & short\_title     & \multicolumn{1}{c|}{}                         &                                \\ \cline{2-2}
\multicolumn{1}{c|}{}                                    & long\_title      & \multicolumn{1}{c|}{}                         &                                \\ \cline{1-2}
\multicolumn{1}{c|}{\multirow{2}{*}{D\_icd\_procedures}} & short\_title     & \multicolumn{1}{c|}{}                         &                                \\ \cline{2-2}
\multicolumn{1}{c|}{}                                    & long\_title      & \multicolumn{1}{c|}{}                         &                                \\ \hline
\end{tabular}%
}\label{table_mapping}
\end{table}
\subsubsection{Labeling Process of MIMIC-OMOP}
MIMIC-OMOP is derived from MIMIC-III, and while both are organized using different table structures (schemas), they contain the same patient information. 
Therefore, the labels of the entities annotated in MIMIC-III can be directly applied to MIMIC-OMOP as well.

We downloaded the MIMIC-OMOP database\footnote{Database sourced from \url{https://github.com/MIT-LCP/mimic-omop}.}, which follows the mapping guidelines specified in Section~\ref{machig_mimic_omop}.
Upon reviewing MIMIC-OMOP, we discovered that the drug exposure start time was recorded as being later than the end time. We corrected this issue by swapping the values in the columns.

\section{Labeling Instructions}
  \label{app: labeling_instruction_rules}
\subsection{Task Description} Identify the entities mentioned in the clinical note and write an SQLite3 query to check if the corresponding values exist in the database. If you encounter any ambiguous or corner cases during the annotation process, make a note of them and discuss them with the practitioners and other annotators.

\subsection{Identify the Entity} 
Identify the entities that likely exist in the following tables: Procedures\_icd, Diagnoses\_icd, Microbiologyevents, Chartevents, Labevents, Prescriptions, Inputevents\_mv, Inputevents\_cv, and Outputevents. Detect entities that occurred exclusively between the patient's admission and the charted time of the note.
\subsubsection{Additional Rules (for Discharge Summary)}
\begin{itemize}[leftmargin=5.5mm]
\item For diagnoses, extract entities from the Discharge Diagnosis list, including both the \textit{Primary Diagnosis} and \textit{Secondary Diagnosis} sections.
\item For procedures, extract entities from both the \textit{Major Surgical} and \textit{Invasive Procedure} sections.
\end{itemize}

\subsection{Classify the Type of the Identified Entity}
\begin{itemize}[leftmargin=5.5mm]
\item Type 1: Entities with numerical values.
\begin{itemize}
    \item When the value associated with the entity is numeric.
    \begin{itemize}
        \item \textit{e.g.}, \textit{temp - 99.9, WBC - 9.6}
    \end{itemize}
\end{itemize}
\item Type 2: Entities without numerical values but whose existence can be verified in the database.
\begin{itemize}
    \item When the value associated with the entity is not numeric and it is sufficient to confirm its existence without checking the value in the database.
    \begin{itemize}
        \item \textit{e.g.}, \textit{Lasix was started, WBC was tested}
    \end{itemize}
\end{itemize}
\item Type 3: Entities with string values.
\begin{itemize}
  \item When the value associated with the entity is not numeric and it is necessary to check the value in the database.
  \begin{itemize}
        \item \textit{e.g.}, \textit{Lasix increased, BP was stable, WBC changed from 1 to 5, temp > 100}
    \end{itemize}
\end{itemize}
\end{itemize}

\subsection{Search Items in the Database}
Use the Item Search Tool to find items in the database related to the detected entity. From these, select those that accurately represent the detected entity.
\begin{itemize}[leftmargin=5.5mm]
\item Guidelines for using the Item Search Tool
\begin{itemize}
    \item Display items related to the entity in the following order: D\_labitems, D\_items, Prescriptions, D\_icd\_procedures, D\_icd\_diagnoses
    \item If none of the searched items can accurately represent the entity, manually search for the items in the database and add it.
\end{itemize}
\end{itemize}

\subsection{Select Tables and Enter the Evidence Line}
\begin{itemize}[leftmargin=5.5mm]
\item Check the tables connected to the selected items and select the tables to which the detected entity can be linked.
\item Enter the line number in the clinical notes where the entity is located.
\end{itemize}

\subsection{Check the Number of Entities}
This section is for checking how many times an entity is mentioned. For example, in the case of BP, if it is written as \textit{120/50}, it should be considered as two mentions of BP. Each instance should then be verified to ensure it is correctly recorded in the database.

\subsection{Extract Information Related to the Entity from the Clinical Note}
\begin{figure}[h]
\includegraphics[width=\linewidth]{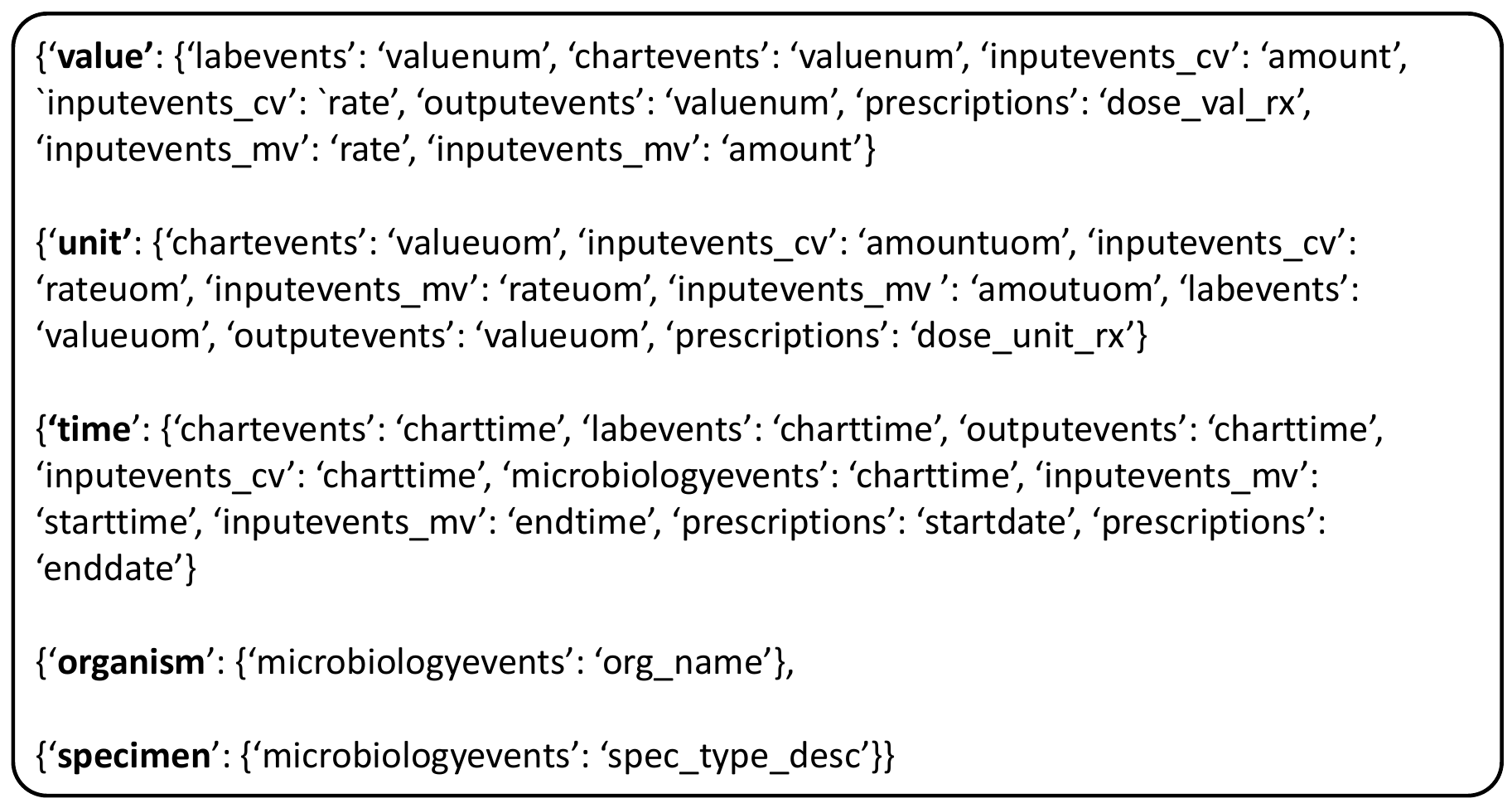}
\vspace{-7mm} 
\caption{Actual table columns corresponding to value, unit, time, organism, and specimen.}
\label{act_col_map}
\end{figure}

Extract the value, unit, time, organism, and specimen related to the entity. Refer to Figure \ref{act_col_map} for the actual table columns corresponding to value, unit, time, organism and specimen.
\begin{itemize}[leftmargin=5.5mm]
\item Value: Numeric value corresponding to the entity (\textit{e.g.}, 184, 103.3). 
\item Unit: Unit corresponding to the numeric value (\textit{e.g.}, mg, ml/h). If the unit is not specified, do not extract it.
\item Time of the clinical event occurrence
\begin{itemize}
    \item If only the date is noted, use the format YYYY-MM-DD. If only MM-DD is noted, use the year from the note's chartdate to complete as YYYY-MM-DD.
    \item In case the time is also specified, use the format YYYY-MM-DD HH:MM:SS (24-hour format).
    \item For relative time expression (\textit{e.g.}, \textit{HD \#3}), calculating the date based on the admission date. For `\textit{Yesterday}', calculate it based on the note's chartdate.
    \item If `admission', `charttime', or `discharge' is noted, calculate the date based on the respective entry.
    \item If there is no time noted, for discharge summaries, consider the entire hospitalization period. For nursing and physician notes, consider the period within one day before and one day after the chartdate.
\end{itemize}
\item Organism: This corresponds to a microbiology event and involves microorganism (\textit{e.g.}, \textit{bacteria, fungi}).
\item Specimen: This corresponds to a microbiology event and involves specimen (\textit{e.g.}, \textit{blood, urine, tissue}) from which the microbiological sample was obtained.
\end{itemize}

\subsection{Example of Annotation}
\begin{itemize}[leftmargin=5.5mm]
\item Physical Exam: Temperature 99.9
\begin{itemize}
    \item Label based on whether there is a record with a value of 99.9 in the `Temperature' data.
\end{itemize}
\item Oxygen saturation 98\%
\begin{itemize}
\item Label based on whether there is a record with a value of 98 and unit of \% in the `Oxygen saturation' data.
\end{itemize}
\item BP: 120/80 (90)
\begin{itemize}
    \item Label whether `BP' has the records of 120, 80, and 90.
\end{itemize}
\end{itemize}

\subsection{Query Generation}
Create an SQL query that utilizes the values extracted from the clinical note and satisfies the conditions specified in Figure \ref{time_expression}. If an inconsistency occurs, mask the condition values, execute the query on the database, and identify the table and columns causing the inconsistency.

\subsection{Notice}
\begin{itemize}[leftmargin=5.5mm]
    \item For blood pressure, if given in the format (num1/num2), each of num1 and num2 should be checked individually.
    \item Exclude entities that are not explicitly identified (\textit{e.g.}, Chem 7: 140 / 4.2 / 104 / 25 / 15 / 1.0 /  90).
\end{itemize}

\section{Time Expressions in Clinical Notes}
\label{app: time_expression}
\begin{figure}[h]
\begin{center}
\includegraphics[width=0.65\linewidth]{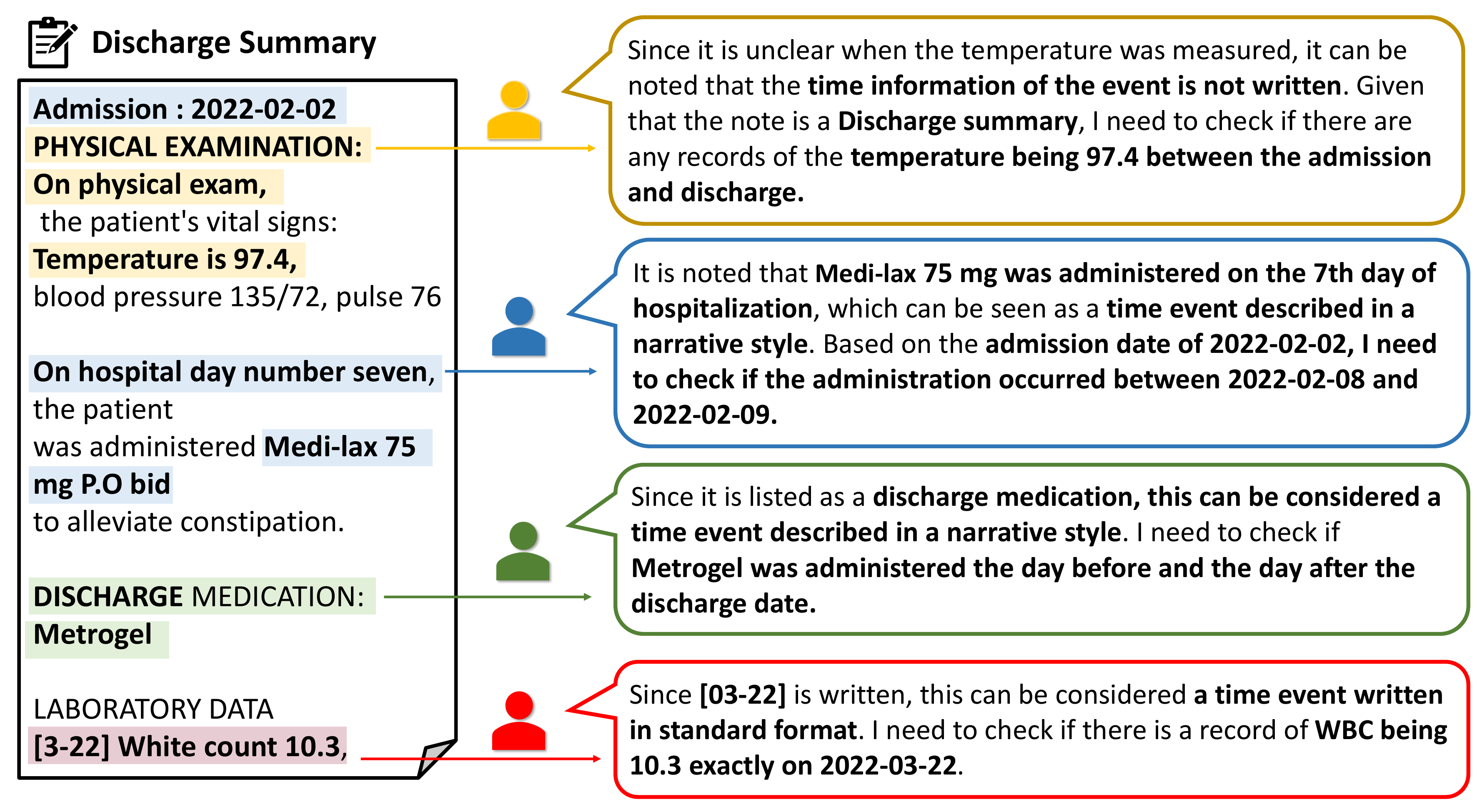}
\vspace{-3mm} 
 \end{center}
\caption{Examples categorized by type of time expression.}
    \label{time_expression2}
 \end{figure}

\begin{figure}[h]
\includegraphics[width=\linewidth]{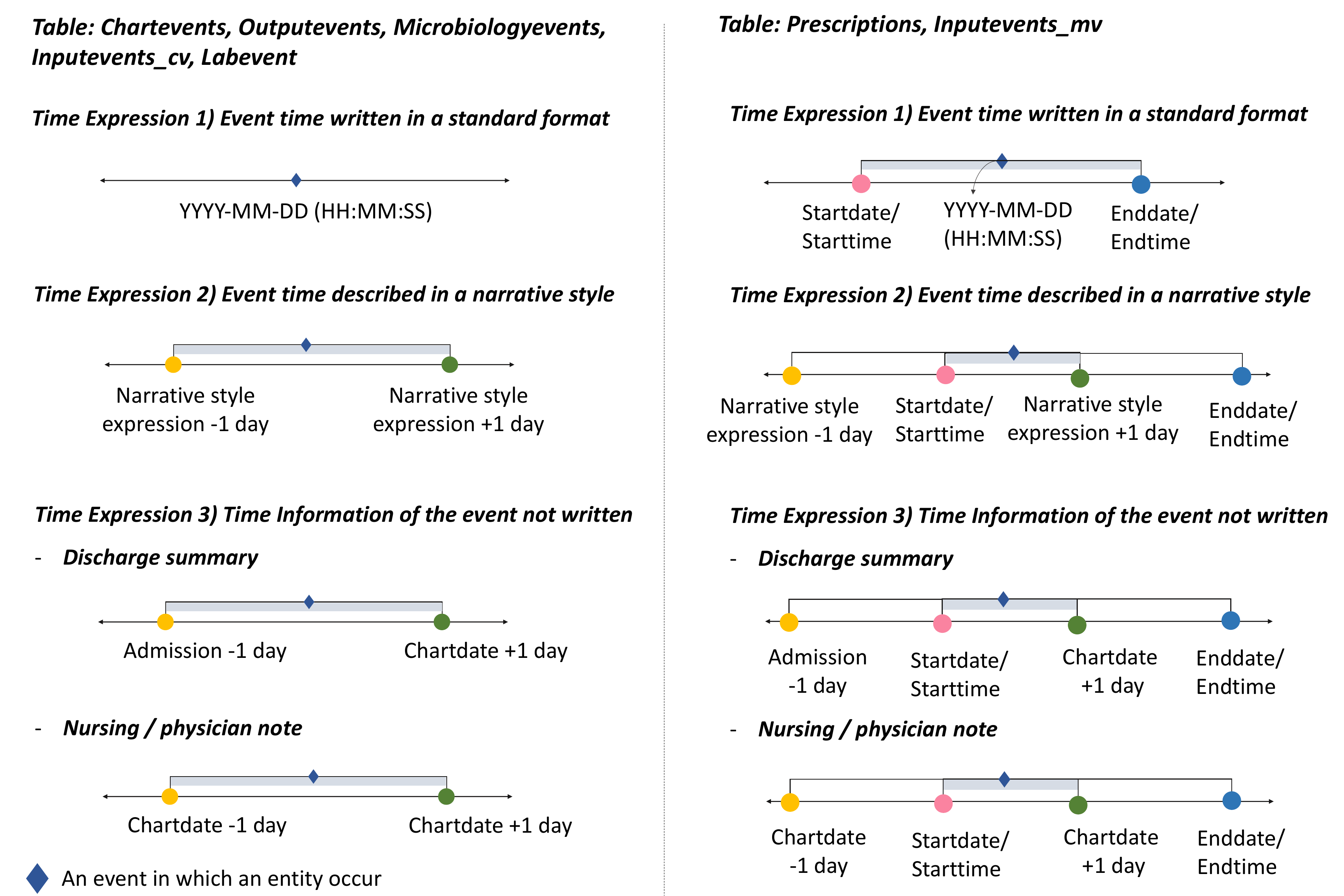}
\vspace{-3mm} 
\caption{Range of verification based on temporal expressions. The left side indicates time solely through `charttime'. On the right, however, `prescriptions' and `inputevents\_mv' utilize both `starttime' and `endtime', creating a varied verification range for time.}
    \label{time_expression}
\end{figure}

Figure~\ref{time_expression2} provides examples categorized by type of time expression. Furthermore, Figure~\ref{time_expression} depicts the verification ranges for various table types based on these time expressions.

\section{Quality Control}
To ensure the high quality of our dataset, we rely on expert researchers for annotation rather than crowd-sourced workers.
The researchers have published AI healthcare-related papers and possess a thorough understanding of the MIMIC-III database and SQL syntax.
After the labeling was completed, four annotators conducted cross-validation on a total of 20 notes.\footnote{Each type of note (\textit{e.g.}, \textit{discharge summary, physician note, nursing note}) has its own unique format and style. Therefore, we manually cross-checked only 20 notes for each type, focusing on these characteristics. For the remaining notes, we corrected the data based on the key elements resolved during the cross-check to maintain consistency.}
As a result, the F1 score of the entities recognized by the annotators was 0.880.
Additionally, among the entities recognized by both annotators, the cases where the labels were the same accounted for 0.938.
This demonstrates the consistency of our labeling process.
\label{app:quality control}

\section{Error cases}
\begin{figure}[h]
    \centering
    \begin{minipage}{0.48\textwidth}
        \centering
     \includegraphics[width=\linewidth]{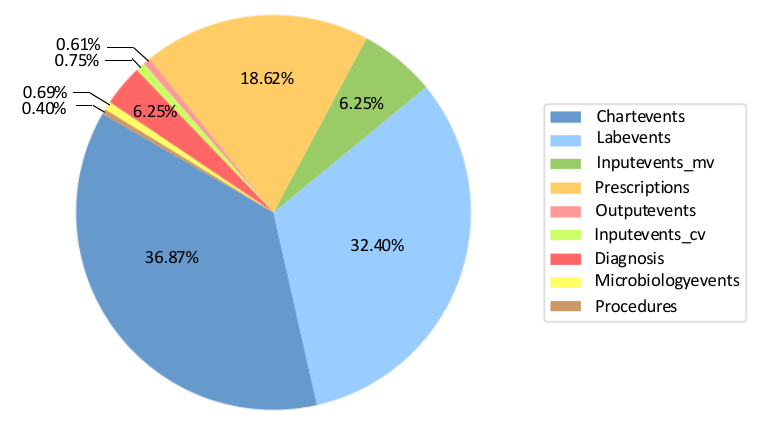}
     \caption{Proportion of tables used in \dname, covering various clinical events in a hospital such as vital signs, medications, and lab results.}
     \label{table_coverage_ablation}
    \end{minipage}
    \hfill
    \begin{minipage}{0.48\textwidth}
        \centering
     \includegraphics[width=\linewidth]{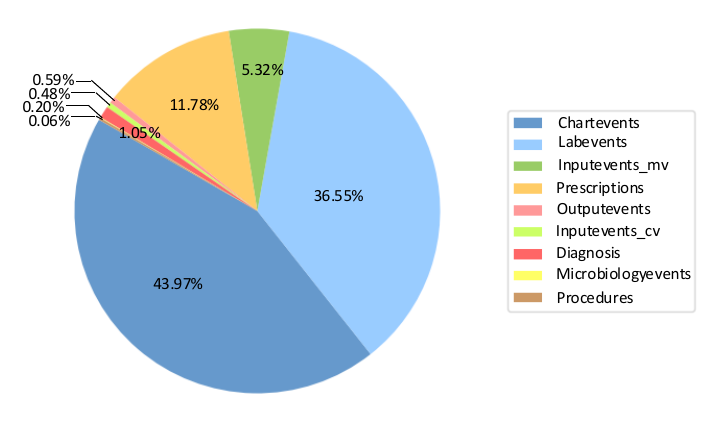}
     \vspace{-5.5mm}
     \caption{Proportion of tables with inconsistencies. Every table exhibited inconsistencies, with the highest rate observed in labevents.}
     \label{table_errors_ablation}
    \end{minipage}
\end{figure}

\begin{figure}[h]
    \centering
    \begin{minipage}{0.42\textwidth}
        \centering
    \includegraphics[width=\linewidth]{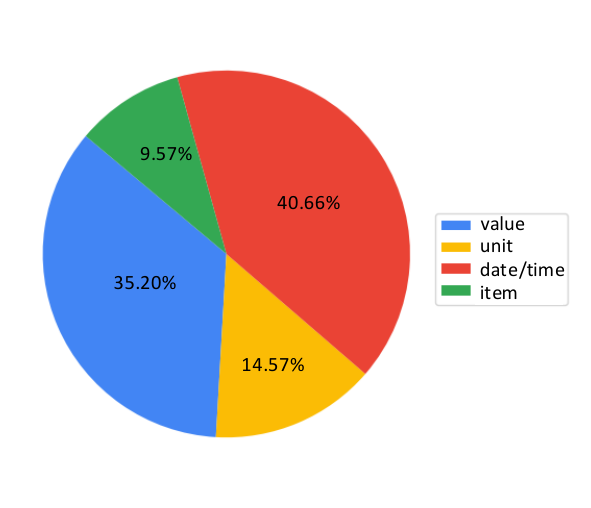}
    \caption{Proportion of columns with inconsistencies. The highest frequency of inconsistencies occurred in the date and time columns.}
    \label{error_cols_ablation}
    \end{minipage}
    \hfill
    \begin{minipage}{0.56\textwidth}
        \centering
    \includegraphics[width=\linewidth]{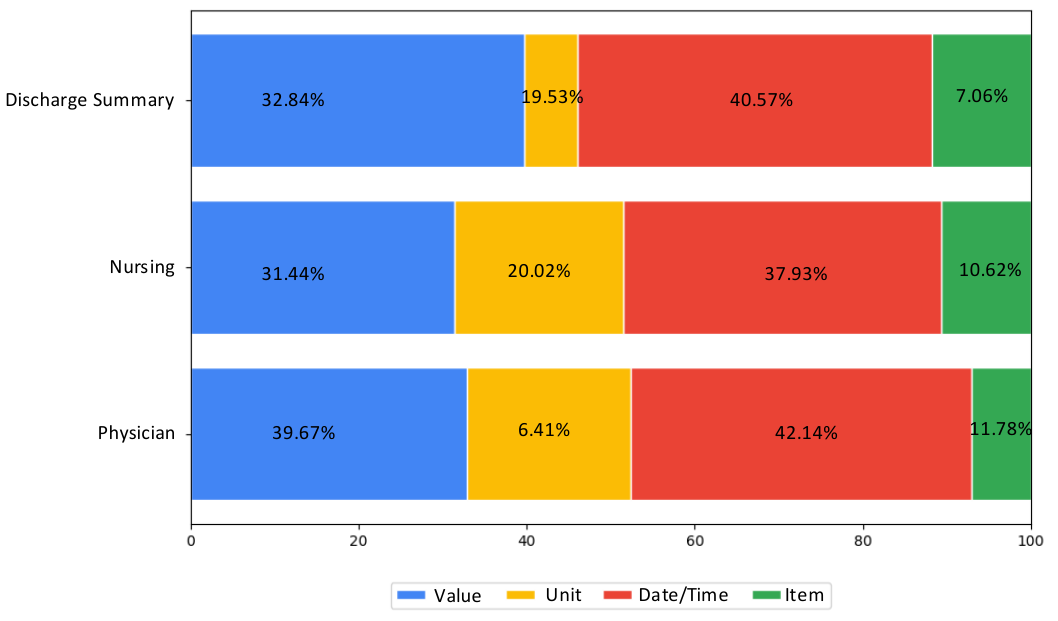}
    \caption{Proportion of inconsistencies by column in each note. Unlike discharge summaries and physician notes, nursing notes had the highest occurrence of inconsistencies within the unit.}
    \vspace{-5mm} 
    \label{column_errors_ablation}
    \end{minipage}
\end{figure}

We conducted an in-depth analysis of various discrepancies observed in \dname. To achieve this, we compared the coverage of tables where discrepancies occur in \dname and analyzed the error rates per column. Additionally, we examined in detail which columns have higher error occurrence rates for each note. These details can be found in Figure \ref{table_coverage_ablation} to \ref{column_errors_ablation}.

\label{app: error_cases}

\section{Prompt}
\label{app: prompt}
The prompts used in \mname can be found in Figure~\ref{prompt_note_seg} to \ref{prompt_query_exact}. To protect patient privacy, all values in the example data mentioned in the paper have been replaced with fictional values. Additionally, some sentences have been paraphrased or omitted.

In particular, table descriptions and column descriptions were utilized to effectively conduct the table identification and pseudo table creation processes. 
The descriptions used for this purpose, derived using MIMIC-III documentation\footnote{https://mimic.mit.edu/docs/iii/} and ChatGPT-4, can be found in Table \ref{desc_table} and Table \ref{desc_column}.

\begin{longtblr}
[
 caption = {Table description used in the Table Identification prompt.},
 label = {desc_table}
]
{
 colspec = {X[1.4,c,m]X[2.2,c,m]X[10,l,m]},
 colsep = 0.3pt,
 rowhead = 1,
 hlines,
 rows={font=\tiny},
 rowsep=0.3pt,
}
\textbf{Database} & \textbf{Table} & {\SetCell[c=1]{c}\textbf{Description}} \\
MIMIC-III & D\_items &  D\_items provides metadata for all recorded items, including medications, procedures, and other clinical measurements, with unique identifiers, labels, and descriptions.\\
MIMIC-III & Chartevents &  Chartevents contains time-stamped clinical data recorded by caregivers, such as vital signs, laboratory results, and other patient observations, with references to the D\_items table for item details.\\
MIMIC-III & Inputevents\_cv &  Inputevents\_cv contains detailed data on all intravenous and fluid inputs for patients during their stay in the ICU and uses ITEMID to link to D\_items. \\
MIMIC-III & Inputevents\_mv &  The inputevents\_mv table records detailed information about medications and other fluids administered to patients, including dosages, timings, and routes of administration, specifically from the MetaVision ICU system.\\
MIMIC-III & Microbiologyevents & Microbiologyevents contains detailed information on microbiology tests, including specimen types, test results, and susceptibility data for pathogens identified in patient samples. This information is linked to D\_items by ITEMID.\\
MIMIC-III & Outputevents &  Records information about fluid outputs from patients, such as urine, blood, and other bodily fluids, including timestamps, amounts, and types of outputs, with references to the D\_items table for item details.\\
MIMIC-III & D\_labitems &  D\_labitems contains metadata about laboratory tests, including unique identifiers, labels, and descriptions for each lab test performed.\\
MIMIC-III & Labevents &  Labevents contains detailed records of laboratory test results, including test values, collection times, and patient identifiers, with references to the D\_labitems table for test-specific metadata.\\
MIMIC-III & Prescriptions &   Lists patient prescriptions with details on dose, administration route, and frequency. There is no reference table.\\
MIMIC-III & D\_icd\_diagnoses &   The D\_icd\_diagnoses table provides descriptions and categorizations for ICD diagnosis codes used to classify patient diagnoses.\\
MIMIC-III & Diagnoses\_icd &  The Diagnoses\_icd table contains records of ICD diagnosis codes assigned to patients, linking each diagnosis to specific hospital admissions.\\
MIMIC-III & D\_icd\_procedures &  D\_icd\_procedures contains definitions and details for ICD procedure codes, including code descriptions and their corresponding categories.\\
MIMIC-III & Procedures\_icd &  Procedures\_icd records the procedures performed on patients during their hospital stay, indexed by ICD procedure codes and linked to specific hospital admissions.\\
MIMIC-OMOP & Concept &  The concept table is a standardized lookup table that contains unique identifiers and descriptions for all clinical and administrative concepts, providing a consistent way to reference data elements across various healthcare domains.\\
MIMIC-OMOP & Measurement &  Table stores quantitative or qualitative data obtained from tests, screenings, or assessments of a patient, including lab results, vital signs, and other measurable parameters related to patient health.\\
MIMIC-OMOP & Durg\_exposure &  Drug\_exposure table records details about the dispensing and administration of drugs to a patient, including the drug type, dosage, route, and duration of each drug exposure event.\\
MIMIC-OMOP & Specimen &  Specimen table contains details about patient specimens, including type, collection method, and collection context.\\
MIMIC-OMOP & Condition\_occurrence &  The table logs instances of clinical conditions diagnosed or reported in a patient, detailing the type of condition, diagnosis date, and source information.\\
MIMIC-OMOP & Procedure\_occurrence &  Procedure\_occurrence table captures details of medical procedures performed on a patient, including the type of procedure, date, and relevant context from the healthcare encounter.\\
\end{longtblr}
\begin{longtblr}
[
 caption = {Column description used in the Pseudo Table Creation prompt.},
 label = {desc_column}
]
{
 colspec = {X[1.4,c,m]X[2,c,m]X[1.7,c,m]X[10,l,m]},
 colsep = 0.3pt,
 rowhead = 1,
 hlines,
 rows={font=\tiny},
 rowsep=0.3pt,
}
\textbf{Database} & \textbf{Table} &\textbf{Column} & {\SetCell[c=1]{c}\textbf{Description}} \\
MIMIC-III & D\_items & Label &  The label column provides a human-readable name for this item. The label could be a description of a clinical observation such as Temperature, blood pressure, and heart rate.\\
MIMIC-III & D\_labitems & Label & The label column provides a human-readable name for this item. The label could be a description of a clinical observation such as wbc, glucose, and PTT.\\
MIMIC-III & D\_icd\_procedures & Short\_title & Short\_title provides a concise description of medical procedures encoded by ICD-9-CM codes.\\
MIMIC-III & D\_icd\_procedures & Long\_title & Long\_title offers a detailed and comprehensive description of medical procedures associated with ICD-9-CM codes. \\
MIMIC-III & D\_icd\_diagnoses & Short\_title & Short\_title provides brief descriptions or names of medical diagnoses corresponding to their ICD-9 codes.\\
MIMIC-III & D\_icd\_diagnoses & Long\_title & Long\_title offers more detailed descriptions of medical diagnoses corresponding to their ICD-9 codes. \\
MIMIC-III & Chartevents & Charttime & Charttime records the time at which an observation occurred and is usually the closest proxy to the time the data was measured, such as admission time or a specific date like 2112-12-12. \\
MIMIC-III & Chartevents & Valuenum & This column contains the numerical value of the laboratory test result, offering a quantifiable measure of the test outcome. If this data is not numeric, Valuenum must be null. \\
MIMIC-III & Chartevents & Valueuom & Valueuom is the unit of measurement. \\
MIMIC-III & Inputevents\_cv & Charttime & Charttime represents the time at which the measurement was charted.\\
MIMIC-III & Inputevents\_cv & Amount & Indicates the total quantity of the input given during the charted event.\\
MIMIC-III & Inputevents\_cv & Amountuom & Amountuom is the unit of Amount.\\
MIMIC-III & Inputevents\_cv & Rate & Details the rate at which the input was administered, typically relevant for intravenous fluids or medications.\\
MIMIC-III & Inputevents\_cv & Rateuom & Rateuom is the unit of Rate.\\
MIMIC-III & Labevents & Charttime & The Charttime column records the exact timestamp when a laboratory test result was charted or documented (\textit{e.g.},\textit{ 2112-12-12)}.\\
MIMIC-III & Labevents & Valuenum & The Valuenum column contains the numeric result of a laboratory test, represented as a floating-point number.\\
MIMIC-III & Labevents & Valueuom & 
The valueuom column  specifies the unit of measurement for the numeric result recorded in the Valuenum column.\\
MIMIC-III & Inputevents\_mv & Starttime & 
The Starttime column records the timestamp indicating when the administration of a medication or other clinical intervention was initiated.\\
MIMIC-III & Inputevents\_mv & Endtime & 
The Endtime column records the timestamp indicating when the administration of a medication or other clinical intervention was completed.\\
MIMIC-III & Inputevents\_mv & Amount & 
Amount records the total quantity of a medication or fluid administered to the patient.\\
MIMIC-III & Inputevents\_mv & Amountuom & 
Amountuom specifies the unit of measurement for the amount of medication or fluid administered, such as milliliters (ml) or milligrams (mg).\\
MIMIC-III & Inputevents\_mv & Rate & 
Rate specifies the rate at which a medication or fluid was administered.\\
MIMIC-III & Inputevents\_mv & Rateuom & 
Rateuom specifies the unit of measurement for the rate at which a medication or fluid was administered, such as milliliters per hour (mL/hr).\\
MIMIC-III & Outputevents & Charttime & 
Charttime records the timestamp when an output event, such as urine output or drainage, was documented.\\
MIMIC-III & Outputevents & Valuenum & 
Valuenum contains the numeric value representing the quantity of output, such as the volume of urine or other fluids, recorded as a floating-point number.\\
MIMIC-III & Outputevents & Valueuom & 
Valueuom specifies the unit of measurement for the numeric value recorded in the Valuenum column, such as milliliters (mL).\\
MIMIC-III & Prescriptions & Startdate & 
The Startdate column records the date when a prescribed medication was first ordered or administered to the patient.\\
MIMIC-III & Prescriptions & Enddate & 
The Enddate column records the date when the administration of a prescribed medication was completed or discontinued.\\
MIMIC-III & Prescriptions & Drug & 
The Drug column lists the name of the medication that was prescribed to the patient.\\
MIMIC-III & Prescriptions & Dose\_val\_rx & 
The Dose\_val\_rx column specifies the numeric value of the prescribed dose for the medication.\\
MIMIC-III & Prescriptions & Dose\_unit\_rx & 
The Dose\_unit\_rx column specifies the unit of measurement for the prescribed dose of the medication, such as milligrams (mg) or milliliters (mL).\\
MIMIC-III & Microbiologyevents & Charttime & 
The Charttime column records the timestamp when the microbiological culture result was documented or charted.\\
MIMIC-III & Microbiologyevents & Org\_name & 
The Org\_name column identifies the name of the organism \textit{(such as a bacterium or fungus)} that was detected in a microbiological culture.\\
MIMIC-III & Microbiologyevents & Spec\_type\_desc & 
The Spec\_type\_desc column describes the type of specimen \textit{(such as blood, urine, or tissue)} from which the microbiological culture was obtained.\\
MIMIC-OMOP & Concept & Concept\_name & 
The Concept\_name column contains an unambiguous, meaningful, and descriptive name for the Concept.\\
MIMIC-OMOP & Drug\_exposure & \makecell{Drug\_exposure\\\_start\_date} & 
The start date for the current instance of Drug utilization. Valid entries include a start date of a prescription, the date a prescription was filled, or the date on which a Drug administration procedure was recorded.\\
MIMIC-OMOP & Drug\_exposure & \makecell{Drug\_exposure\\\_end\_date} & 
The end date for the current instance of Drug utilization. It is not available from all sources.\\
MIMIC-OMOP & Drug\_exposure & Quantity & 
The quantity column records the amount of the drug administered or prescribed, providing essential information for dosage and treatment analysis.\\
MIMIC-OMOP & Drug\_exposure & \makecell{Dose\_unit\\\_source\_value} & 
The dose\_unit\_source\_value captures the original unit of measurement for the drug dosage as recorded in the source data, preserving the raw data detail for reference and mapping purposes.\\
MIMIC-OMOP & Measurements & \makecell{Measurement\\\_datetime} & 
The measurement\_datetime records the exact date and time when the measurement was taken, providing precise temporal context for each measurement entry.\\
MIMIC-OMOP & Measurements & \makecell{Value\_as\\\_number} & 
The value\_as\_number stores the numerical result of the measurement, allowing for quantitative analysis of the recorded data.\\
MIMIC-OMOP & Measurements & \makecell{Unit\_source\\\_value} & 
The unit\_source\_value captures the original unit of measurement as recorded in the source data, preserving the context of the measurement's unit before any standardization.\\
MIMIC-OMOP & Specimen & \makecell{Specimen\\\_datetime} & 
The specimen\_datetime records the exact date and time when the specimen was collected, providing precise temporal context for each specimen entry.\\

\end{longtblr}

\newpage
\begin{figure}[H]
\centering
\input{app_fig/prompt_note_seg}
\caption{Prompt Template for Note Segmentation.}
\label{prompt_note_seg}
\end{figure}

\newpage
\begin{figure}[H]
\centering
\input{app_fig/prompt_ner}
\caption{Prompt Template for Named Entity Recognition.}
\label{prompt_ner}
\end{figure}

\newpage
\begin{figure}[H]
\centering
\input{app_fig/prompt_time}
\caption{Prompt Template for Time Filtering.}
\label{prompt_time}
\end{figure}

\newpage
\begin{figure}[H]
\centering
\input{app_fig/prompt_table_iden}
\caption{Prompt Template for Table Identification.}
\label{prompt_table_iden}
\end{figure}

\newpage
\begin{figure}[H]
\centering
\input{app_fig/prompt_pseudo_cr}
\caption{Prompt Template for Pseudo Table Creation. The example is a prescriptions table.}
\label{prompt_pseudo_cr}
\end{figure}

\newpage
\begin{figure}[H]
\centering
\input{app_fig/prompt_self_corr}
\caption{Prompt Template for Self Correction.}
\label{prompt_self_corr}
\end{figure}

\newpage
\begin{figure}[H]
\centering
\input{app_fig/prompt_reformat}
\caption{Prompt Template for Value Reformatting. The example is a Chartevents table.}
\label{prompt_reformat}
\end{figure}

\newpage
\begin{figure}[H]
\centering
\input{app_fig/prompt_query_exact}
\caption{Prompt Template for Query Generation. This is an example where the date is written in the yyyy-mm-dd (hh:mm:ss) format.}
\label{prompt_query_exact}
\end{figure}

\newpage
\section{Note Segmentation}
\label{app: algorithm}
\begin{figure}[h]
\includegraphics[width=\linewidth]{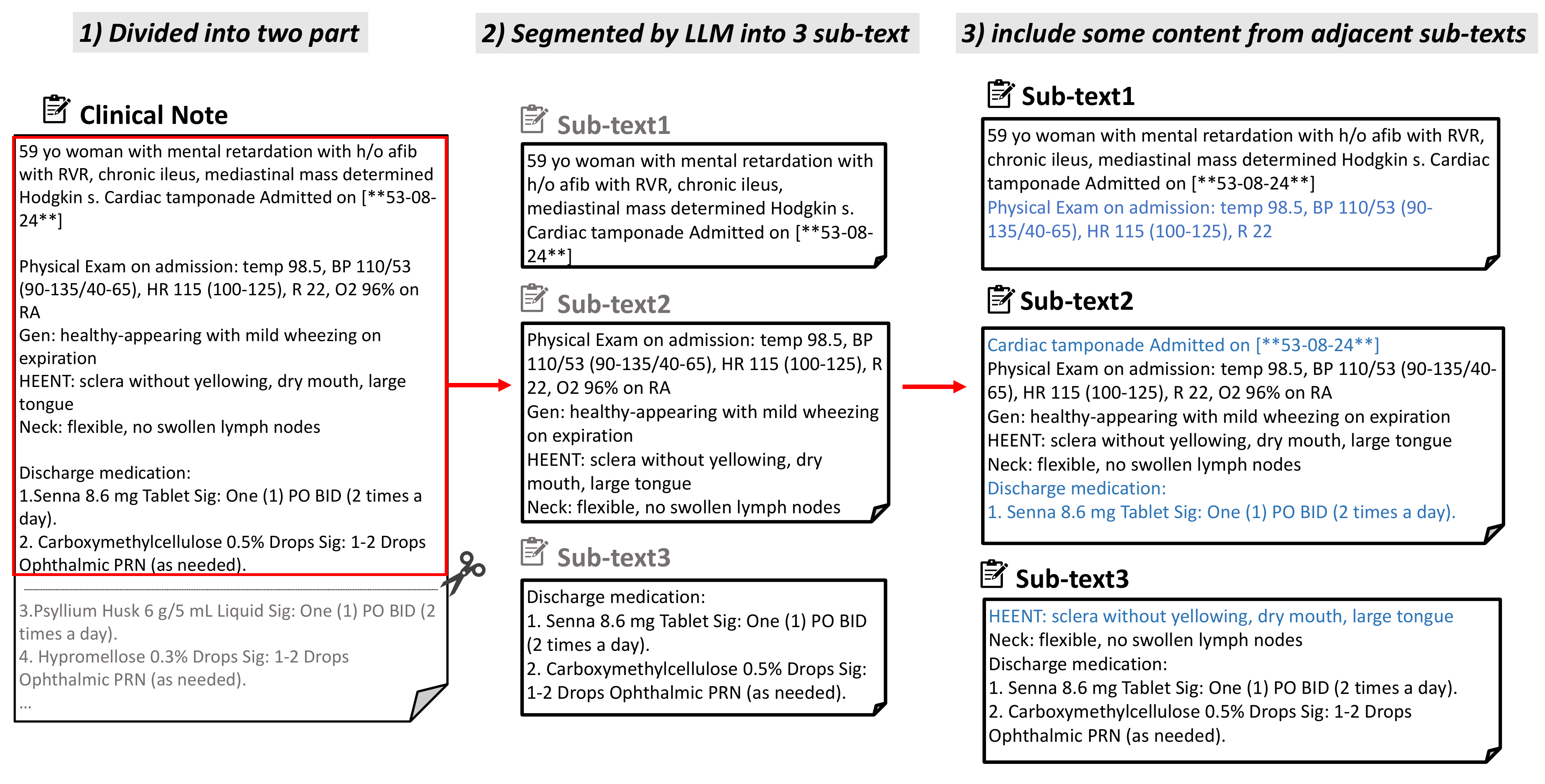}
\caption{Overall process of Note Segmentation where $n$ is 3.}
    \label{note_seg_prosc_fig}
 \end{figure}

Algorithm~\ref{alg:findmax} provides a summary of the note segmentation process, with a detailed example illustrated in Figure~\ref{note_seg_prosc_fig}.

\begin{algorithm}[H]
\centering
\caption{Note Segmentation Process}
\label{alg:findmax}
\begin{algorithmic}[1]
\REQUIRE Clinical Note $P$, Maximum Length of Subtext $l$, Number of Subtexts $n$
\ENSURE Set of Subtexts $\mathcal{T}$
\STATE $\mathcal{T} \leftarrow \emptyset, i \leftarrow 0, P_i \leftarrow P$
\WHILE{$TokenLen(P_i)$ $ > l$}
    \STATE $P^f_i, P^b_i \leftarrow DivideByLen(P_i, l)$
    \STATE $P^f_{i, 1}, P^f_{i, 2}, ..., P^f_{i, n},  \leftarrow DivideByLLM(P^f_i, n)$
    \STATE $\mathcal{T} \leftarrow \mathcal{T} \cup \{P^f_{i, 1}, P^f_{i, 2}, ...P^f_{i, n-1}\}$
    \STATE $P_{i+1} \leftarrow MergeText(P^f_{i, n}, P^b_i)$
    \STATE $i \leftarrow i+1$    
\ENDWHILE
\STATE $\mathcal{T} \leftarrow \mathcal{T} \cup \{P_{i}\}$
\STATE $\mathcal{T} \leftarrow MakeIntersection(\mathcal{T})$
\RETURN $\mathcal{T}$
\end{algorithmic}
\end{algorithm}

\subsection{Effectiveness of Note Segmentation}
Recent advancements in LLMs with extended context lengths have introduced the possibility of processing entire clinical notes without segmentation, potentially capturing more contextual information for error detection tasks. However, upon evaluating this approach on 25\% of our test set, we found that models processing unsegmented notes consistently showed a significant decrease in recall compared to their segmented counterparts—as seen in Table~\ref{noteseg_tab}, while there was a slight increase in precision for some models, the overall ability to detect errors diminished. GPT-3.5 with a 16k context window (without segmentation) achieved a recall of 48.64\% and precision of 54.53\%, whereas with segmentation (using the 4k context window), it achieved a higher recall of 65.45\% and a precision of 50.75\%. This suggests that current long-context LLMs may struggle with accurately extracting detailed information from extensive free-form clinical notes, leading to missed errors. Therefore, our note segmentation approach remains effective, as it enhances the models' ability to detect errors by focusing on relevant sections and allows the use of models with shorter context lengths, which are more accessible and practical for deployment across different healthcare settings.

\begin{table}[htbp]
\caption{Experiment Results for Note Segmentation}
\label{noteseg_tab}
\centering
\renewcommand{\arraystretch}{1.5} 
\begin{tabular}{c c c c}
    \hline
    \multicolumn{1}{c}{\textbf{Experiment Setting}} & \multicolumn{1}{c}{\textbf{Model}} & \textbf{Recall} & \textbf{Precision} \\ 
    \hline
    \multirow{3}{*}{w/ Note Segmentation} & GPT-3.5 4k (0613) & 65.45 & 50.75 \\ 
    \cline{2-4}
    & Llama 3 70B  & 53.86 & 47.01 \\ 
    \cline{2-4}
    & Mixtral 8X7B & 54.92 & 44.60 \\ 
    \hline
    \multirow{3}{*}{w/o Note Segmentation} & GPT-3.5 16k (0613) & 48.64 & 54.53 \\ 
    \cline{2-4}
    & Llama 3.1 70B & 35.23 & 45.40 \\ 
    \cline{2-4}
    & Mixtral 8X7B & 36.43 & 42.58 \\ 
    \hline
\end{tabular}
\end{table}


\section{Process of Creating Pseudo Table}
\label{app: pseudo table}
\begin{figure}[h]
\includegraphics[width=\linewidth]{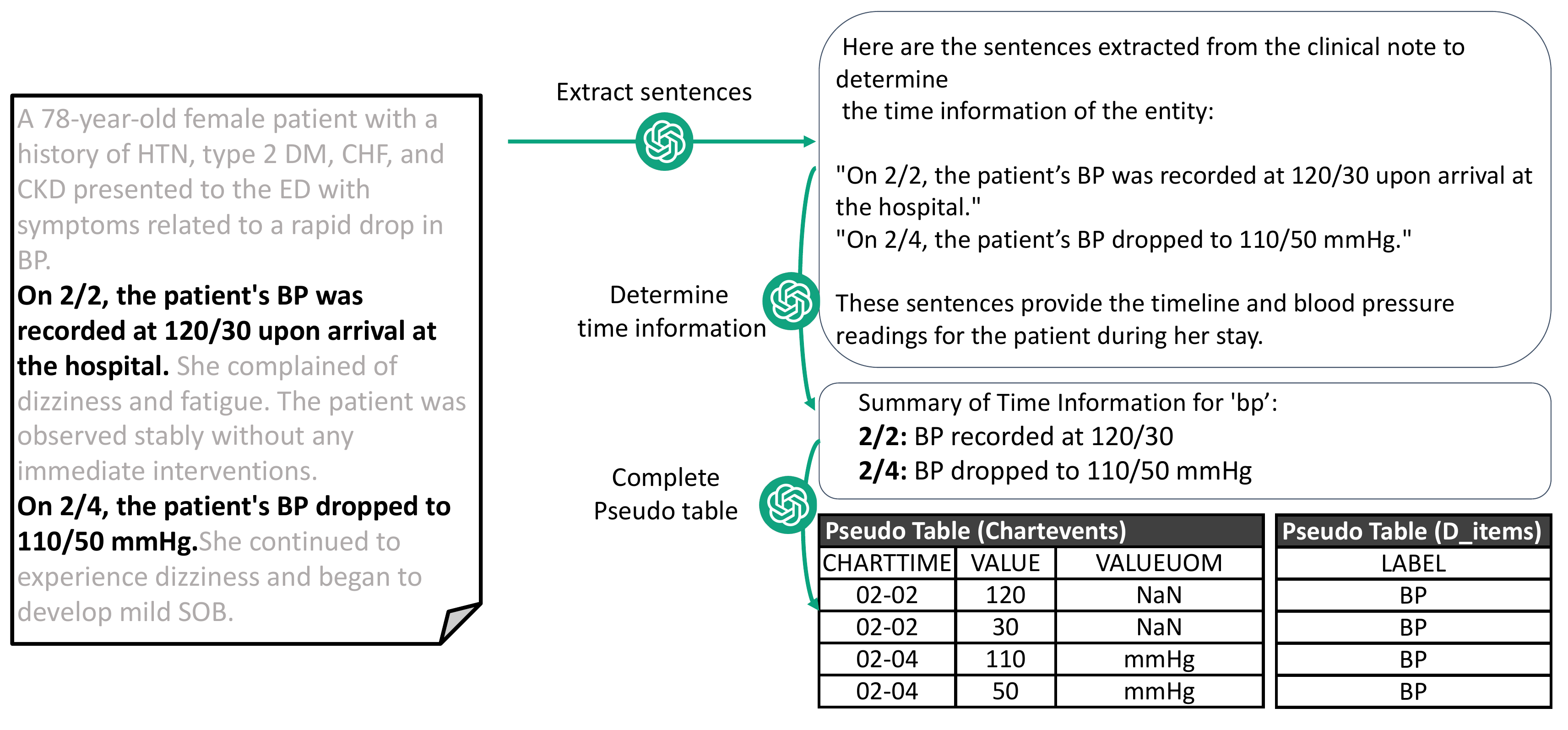}
\caption{The overall process of creating a pseudo table. First, the LLM identifies sentences about a given entity (\textit{BP}) in the text. Then, it finds time information in those sentences and uses this information to complete the pseudo table.}
    \label{pseudo_stat}
 \end{figure}

An example of the process for creating a pseudo table is described in Figure~\ref{pseudo_stat}.

\section{Hallucination of LLMs}
\label{app: hallucination}
\begin{figure}[h]
\begin{center}
\includegraphics[width=0.5\linewidth]{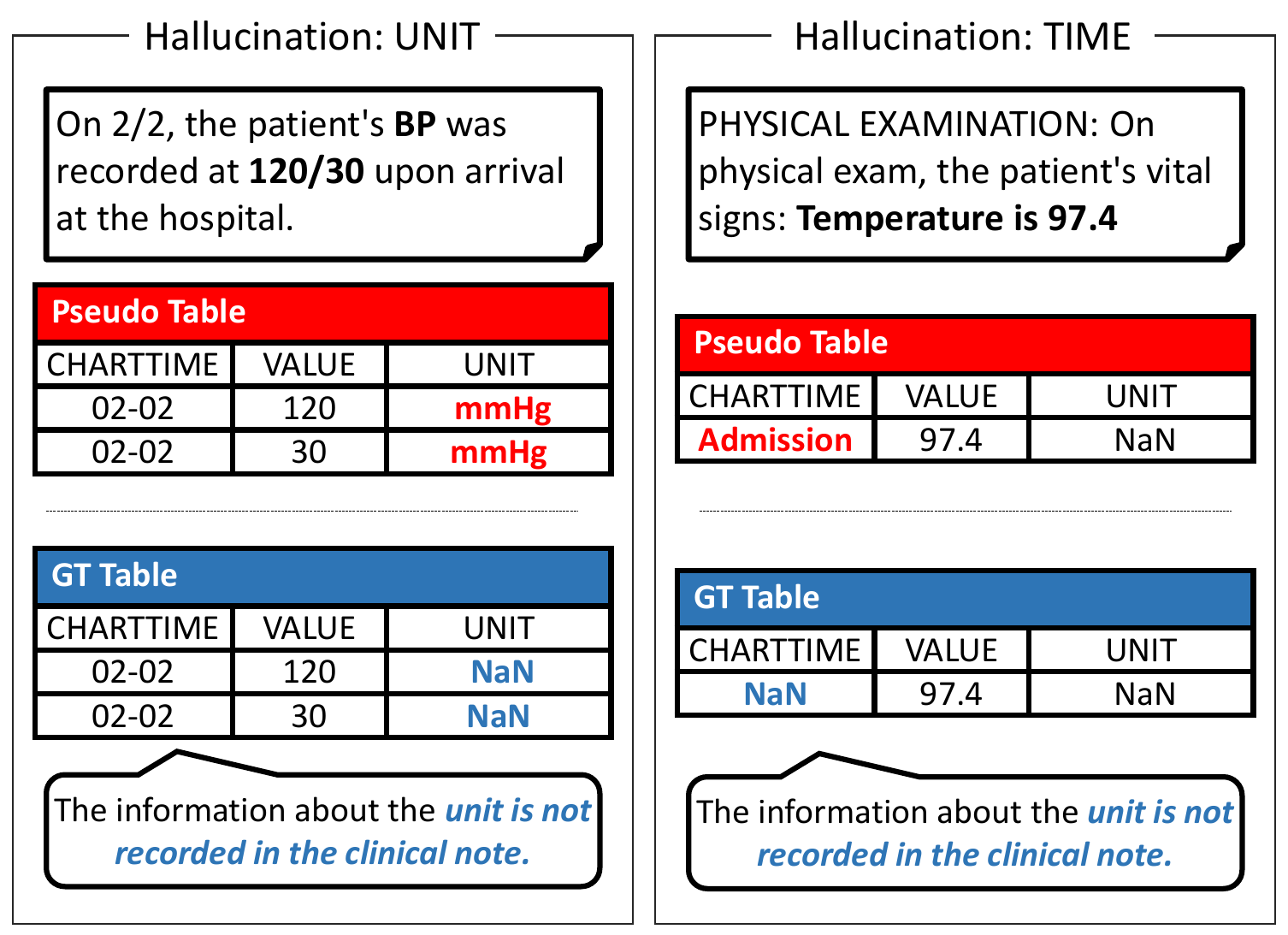}
 \end{center}
\caption{Examples of hallucination in LLMs. 
There have been many instances where the LLM generated non-existent information from clinical notes to create pseudo tables.
In the left box, a pseudo table was created with the unit listed as `mmHg', which was not mentioned in the note. In the right box, a pseudo table was created with the charttime listed as `admission', based on the `PHYSICAL EXAMINATION' section, which was not explicitly mentioned.}
    \label{hallucination_example}
 \end{figure}

Examples of hallucination in LLMs are described in Figure~\ref{hallucination_example}.

\section{Experiments using Original Notes}
 \label{app: origianl_note_exp}
The results using the unfiltered original notes are reported in Table~\ref{org}. Clinical notes often contain not only information related to the current hospitalization but also past medical history or future plans, which can be difficult to discern from tables. To effectively filter this information, we use a Time Filtering step where an LLM determines if the entity occurred during the current hospitalization. We proceed with further steps (\textit{e.g.}, pseudo table creation) only if the LLM determines that the entity is relevant to the current hospitalization.
The experimental results showed that Recall and Precision decreased by approximately 12.73\% and 8.31\%, respectively, compared to the filtered notes. This indicates that the current LLMs lack the reasoning ability to understand clinical notes and determine whether each event occurred during the current admission.

\renewcommand{\arraystretch}{1.4}
\begin{table}[]
\caption{Results using the unfiltered original notes. We conduct experiments using a few-shot setting.}
\resizebox{\columnwidth}{!}{%
\begin{tabular}{cccclccclccclccc}
\hline
\multirow{2}{*}{\textbf{Models}} & \multicolumn{3}{c}{\textbf{Discharge Summary}} & \textbf{} & \multicolumn{3}{c}{\textbf{Physician Note}} & \textbf{} & \multicolumn{3}{c}{\textbf{Nursing Note}} & \textbf{} & \multicolumn{3}{c}{\textbf{Total}} \\ \cline{2-4} \cline{6-8} \cline{10-12} \cline{14-16} 
 & Rec & Prec & \multicolumn{1}{l}{Inters} &  & Rec & Prec & \multicolumn{1}{l}{Inters} &  & Rec & Prec & \multicolumn{1}{l}{Inters} &  & Rec & Prec & \multicolumn{1}{l}{Inters} \\ \hline
Tulu2 & 31.90 & 40.75 & 69.18 &  & 39.78 & 43.43 & 85.64 &  & 39.07 & 29.10 & 68.56 &  & 36.91 & 37.76 & 74.46 \\
Mixtral & 39.16 & 37.32 & 72.52 &  & 37.55 & 40.86 & 99.11 &  & 47.86 & 31.38 & 70.08 &  & 41.52 & 36.52 & 81.57 \\
Llama-3 & 41.91 & 36.88 & 69.34 &  & 37.2 & 40.44 & 79.37 &  & 32.09 & 27.20 & 54.23 &  & 37.06 & 34.84 & 67.64 \\ \hline
\end{tabular}%
}\label{org}
\end{table}

\section{Experiments Setting Details}
\label{app:evaluate metrics example}
\subsection{Examples of Evaluation Metrics}
We measured the performance of the framework using the following three metrics: Recall, Precision, and Intersection. We demonstrate how these metrics are calculated with the following examples.
Consider the gold entity set $\mathcal{G} = \{ \{ e_1, i \}, \{ e_2, c \}, \{ e_3, c \} \}$, and the recognized entity set $\mathcal{R} = \{ \{ e_1, i \}, \{ e_3, i \}, \{ e_4, c \} \}$, where $e_n$ represents an entity, $i$ indicates \textsc{Inconsistent}, and $c$ indicates \textsc{Consistent}. In this situation,  the Recall is 33.33\% because only 1 out of the 3 labeled entities, $\{ e_1, i \}$, was correctly classified. Similarly, the Precision is 33.33\% since only 1 out of the 3 recognized entities, $\{e_1, i\}$, was accurate. However, the Intersection is 50.00\% because the intersection of the gold and recognized sets includes 2 entities, $\{e_1, i\}$ and $\{e_3, i\}$, and only 1 of these, $\{e_1, i\}$, was correctly classified.

\subsection{Number of In-Context Examples}
We carefully designed in-context examples for each stage to maximize the performance of the framework. We provided 15 examples for Table Identification stage and 2 examples for all other stages.

\section{Results of MIMIC-III and MIMIC-OMOP}
\label{app:type exp}
Table~\ref{note_entity_type} presents the experimental results categorized by entity type and note type from both MIMIC-III and MIMIC-OMOP.

\begin{table}[h]
\centering
\caption{The results of MIMIC-III and MIMIC-OMOP. This performance metric is Recall.}
\resizebox{0.8\columnwidth}{!}{%
\begin{tabular}{c|c|c|c|c|c|c}
\hline
\textbf{Data} & \textbf{Type} & \textbf{Model} & \textbf{Discharge Summary} & \textbf{Physician Note} & \textbf{Nursing Note} & \textbf{Total} \\ \hline
\multicolumn{1}{c|}{\multirow{6}{*}{\textbf{MIMIC-III}}} & \multirow{3}{*}{\textbf{Type 1}} & Tulu2 & 69.26 & 73.38 & 68.75 & 66.79 \\
\multicolumn{1}{c|}{} &  & Mixtral & 68.70 & 75.50 & 68.93 & 59.21 \\
\multicolumn{1}{c|}{} &  & Llama3 & 65.15 & 72.88 & 59.21 & 65.74 \\ \cline{2-7} 
\multicolumn{1}{c|}{} & \multirow{3}{*}{\textbf{Type 2}} & Tulu2 & 19.56 & 43.06 & 40.34 & 34.32 \\
\multicolumn{1}{c|}{} &  & Mixtral & 19.49 & 47.38 & 49.47 & 38.31 \\
\multicolumn{1}{c|}{} &  & Llama3 & 20.10 & 42.76 & 56.68 & 39.84 \\ \hline
\multicolumn{1}{c|}{\multirow{6}{*}{\textbf{MIMIC-OMOP}}} & \multirow{3}{*}{\textbf{Type 1}} & Tulu2 & 54.93 & 52.41 & 54.75 & 54.03 \\
\multicolumn{1}{c|}{} &  & Mixtral & 51.88 & 57.34 & 47.98 & 52.40 \\
\multicolumn{1}{c|}{} &  & Llama3 & 51.17 & 53.35 & 53.08 & 52.53 \\ \cline{2-7} 
\multicolumn{1}{c|}{} & \multirow{3}{*}{\textbf{Type 2}} & Tulu2 & 49.77 & 47.13 & 47.05 & 47.98 \\
\multicolumn{1}{c|}{} &  & Mixtral & 50.50 & 35.41 & 50.41 & 45.44 \\
\multicolumn{1}{c|}{} &  & Llama3 & 54.16 & 53.23 & 49.85 & 52.41 \\ \hline
\end{tabular}%
}
\label{note_entity_type}
\end{table}

\section{Component Analysis of \mname}
\label{app: component analysis}
All main experiments utilized the Item Search Tool (see Sec. \ref{sec3.2}) to search for items related to entities in the database. However, the actual annotated data also includes instances where annotators manually searched for items in the database when the search tool did not yield results. As seen in Table \ref{item_search_ab}, the experiment showed that using both the outputs from the Item Search Tool and the additional manual annotations by annotators resulted in an increase in Recall and Precision by 1.75\% and 4.85\%, respectively, compared to using only the Item Search Tool. Therefore, future work should explore methods to find semantically similar items in the database, rather than relying solely on the surface form of the entity.

Table~\ref{app: model_other1} shows the results of experiments for component analysis conducted by matching entities with items in the database, including items added by the Item Search Tool and annotators. 
The open-source models also demonstrated an average performance improvement of 10\% in each setting, proving that each stage plays a crucial role in solving this task. 
Unlike other models that showed significant performance improvements in the second and third settings, Llama3's Recall remained at 74.96\% in the third setting. 
This indicates that Llama3's SQL query generation capability is inferior to that of other models. 
By understanding the performance of each LLM in different settings and addressing their shortcomings, the framework's performance will be significantly enhanced.


\begin{table}[h]
\caption{Component Analysis Result}

\resizebox{\columnwidth}{!}{%
\begin{tabular}{c|l|cc|c|l|lc}
\hline
\textbf{Models} & \multicolumn{1}{c|}{\textbf{Experiment Setting}} & \textbf{Rec} & \textbf{Prec} & \textbf{Models} & \multicolumn{1}{c|}{\textbf{Experiment Setting}} & \multicolumn{1}{c}{\textbf{Rec}} & \textbf{Prec} \\ \hline
\multirow{4}{*}{\textbf{Tulu2}} & CheckEHR & 50.69 & 45.93 & \multirow{4}{*}{\textbf{Llama3}} & CheckEHR & \multicolumn{1}{c}{53.86} & 41.57 \\
 & - NER & 61.86 & 74.27 &  & - NER & 60.33 & 64.67 \\
 & - (Table Identification + Time Filtering) & 67.53 & 69.69 &  & - (Table Identification + Time Filtering) & 67.12 & 68.81 \\
 & - Pseudo Table Creation & 90.35 & 93.27 &  & - Pseudo Table Creation & 74.96 & 87.32 \\ \hline
\multirow{4}{*}{\textbf{Mixtral}} & CheckEHR & 54.92 & 44.60 & \multirow{4}{*}{\textbf{GPT-3.5 (0613)}} & CheckEHR & \multicolumn{1}{c}{65.45} & 50.75 \\
 & - NER & 66.35 & 70.51 &  & - NER & 76.11 & 77.70 \\
 & - (Table Identification + Time Filtering) & 70.05 & 74.86 &  & - (Table Identification + Time Filtering) & 82.49 & 78.99 \\
 & - Pseudo Table Creation & 86.06 & 91.26 &  & - Pseudo Table Creation & 92.83 & 94.93 \\ \hline
\end{tabular}%
}
\label{app: model_other1}
\end{table}

\begin{table}[h]
\centering
\caption{Results of experiments for component analysis conducted by matching entities with items in the database. 
In our experiment, we bypassed the named entity recognition (NER) step and directly provided gold entities. -Additional Manual Annotation uses only the Item Search Tool, whereas +Additional Manual Annotation uses both the Item Search Tool and additional manual annotation.}
\resizebox{0.7\columnwidth}{!}{%
\begin{tabular}{c|cc|cc}
\hline
\multirow{2}{*}{\textbf{Model}} & \multicolumn{2}{c|}{\textbf{\begin{tabular}[c]{@{}c@{}}-Additional \\ Manual Annotation\end{tabular}}} & \multicolumn{2}{c}{\textbf{\begin{tabular}[c]{@{}c@{}}+Additional \\ Manual Annotation\end{tabular}}} \\ \cline{2-5} 
 & \textbf{Rec} & \textbf{Prec} & \textbf{Rec} & \textbf{Prec} \\ \hline
\textbf{Tulu2} & 61.56 & 65.07 & 61.86 & 74.27 \\
\textbf{Mixtral} & 65.07 & 70.21 & 66.35 & 70.51 \\
\textbf{Llama3} & 59.23 & 56.16 & 60.33 & 64.67 \\
\textbf{GPT-3.5 (0613)} & 71.81 & 76.25 & 76.11 & 77.70 \\ \hline
\end{tabular}%
}\label{item_search_ab}
\end{table}

\newpage
\section{Sample data of \dname}
\begin{figure}[h]
\includegraphics[width=\linewidth]{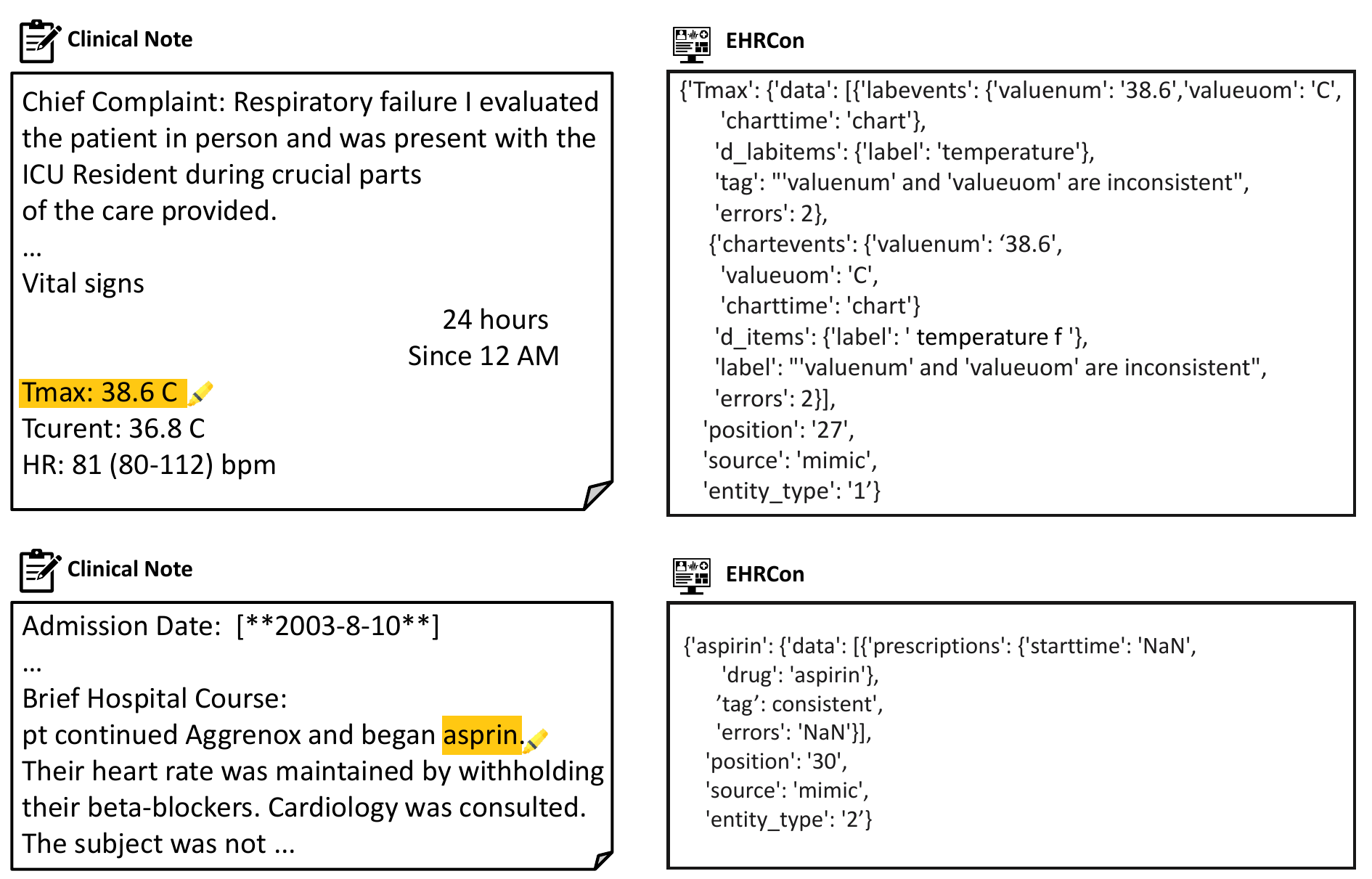}
\caption{These are sample datas of \dname. The `Tmax' example illustrates an inconsistent case, while the `aspirin' example demonstrates a consistent case. The `tag' provides the results of the verification, the `errors' indicate the number of errors, and the `position' refers to the sentence number containing each entity. The `entity\_type' of `1' indicates that a numeric value is clearly shown, while the `entity\_type' of `2' means that the presence of the entity in the database is sufficient for verification. If the `time' is marked as `NaN', it means there is no time information associated with the entity. In the provided samples, all entities and values have been altered to prevent patient identification, and the sentences within the clinical notes have been rephrased with some content added or removed.}

    \label{ehrcon_sample}
 \end{figure}

The sample data of \dname is described in Figure~\ref{ehrcon_sample}.

 \newpage
\section{SQL Generation Method}
\label{sql_gen_methodology}
In our data annotation process, we adopted a template-based method for generating SQL queries to ensure both consistency and precision. Annotators filled in predefined SQL templates, with specific slots for table names, column names, patient identifiers, time expressions, and condition values \textit{(e.g., SELECT * FROM} \{table1\}
\textit{JOIN} \{table2\} \textit{ON} \{table1\}\textit{.}\{column\_name1\} \textit{=} \{table2\}\textit{.}\{column\_name2\}
\textit{WHERE} \{table1\}\textit{.}hadm\_id \textit{=} \{hadm\_id\} \textit{AND} [time\_value\_template] \textit{AND} \{table1\}\textit{.}[condition\_value\_column] \textit{=} \{condition\_value\}).

To cover a range of time expressions found in clinical notes, the [time\_value\_template] accommodated not only standard formats such as `\textit{YYYY-MM-DD HH:MM:SS}' but also narrative descriptors like `\textit{admission}' or `\textit{discharge}'. It even handled cases where time information was entirely absent, allowing for flexibility in the SQL query generation process.
As a result, 27 unique templates were developed, all recorded in Table~\ref{sql_tempalte}, and carefully reviewed and validated by two authors. This validation process included simulating potential queries and cross-checking the results against the underlying database to ensure the templates accurately captured the intended data retrieval logic.

Annotators relied on these validated templates to generate SQL queries throughout the annotation task, which significantly contributed to maintaining high reliability in data extraction. This approach's effectiveness was further supported by a high inter-annotator agreement rate of 93.8\%, as seen in cross-checked annotations (refer to Appendix~\ref{app:quality control}). This consistency in the annotations indirectly validated the accuracy of the generated SQL queries, reinforcing the robustness of our approach.

\begin{longtblr}
[
 caption = {SQL template used for data annotation},
 label = {sql_tempalte}
]
{
 colspec = {X[4,c,m]X[24,l,m]},
 colsep = 0.3pt,
 rowhead = 1,
 hlines,
 rows={font=\tiny},
 rowsep=0.3pt,
}
\textbf{Table} & {\SetCell[c=1]{c}\textbf{SQL template}} \\
\SetCell[r=4]{c} Chartevents  & SELECT * FROM Chartevents JOIN d\_items ON Chartevents.itemid d\_items.itemid WHERE strftime(`\%Y-\%m-\%d’, Chartevents.charttime) = `\{date\_value\}’ AND \{condition\_values\};\\
 & SELECT *FROM Chartevents JOIN d\_items ON Chartevents.itemid = d\_items.itemid WHERE strftime(`\%Y-\%m-\%d’, Chartevents.charttime) BETWEEN strftime(`\%Y-\%m-\%d’, datetime(`\{admission\}’, `-1 day’)) AND strftime(`\%Y-\%m-\%d’, datetime(`\{admission\}’, `+1 day’))AND \{condition\_value\};\\
 & SELECT * FROM Chartevents JOIN d\_items ON Chartevents.itemid = d\_items.itemid WHERE strftime(`\%Y-\%m-\%d’, Chartevents.charttime) BETWEEN strftime(`\%Y-\%m-\%d’, datetime(`\{admission\}’, `-1 day’)) AND strftime(`\%Y-\%m-\%d’, datetime(`\{chart\}’, `+1 day’)) AND \{condition\_value\};\\
 & SELECT * FROM Chartevents JOIN d\_items ON Chartevents.itemid = d\_items.itemid WHERE strftime(`\%Y-\%m-\%d’, Chartevents.charttime) BETWEEN strftime(`\%Y-\%m-\%d’, datetime(`\{calculated\_time\}’, `-1 day’)) AND strftime(`\%Y-\%m-\%d’, datetime(`\{calculated\_time\}’, `+1 day’)) AND \{condition\_value\};\\ 
\SetCell[r=4]{c} Labevents  & SELECT * FROM Labevents JOIN d\_labitems ON Labevents.itemid = d\_labitems.itemid WHERE strftime(`\%Y-\%m-\%d’,
Labevents.charttime) = `\{date\_value\}’ AND \{condition\_values\};\\
 & SELECT * FROM Labevents JOIN d\_labitems ON Labevents.itemid = d\_labitems.itemid WHERE strftime(`\%Y-\%m-\%d’,chartevents.charttime) BETWEEN strftime(`\%Y-\%m-\%d’, datetime(`{admission}’, `-1 day’)) AND strftime(`\%Y-\%m-\%d’, datetime(`{admission}’, `+1 day’))AND \{condition\_value\};\\
 & SELECT * FROM Labevents JOIN d\_labitems ON Labevents.itemid = d\_labitems.itemid WHERE strftime(`\%Y-\%m-\%d’, Labevents.charttime) BETWEEN strftime(`\%Y-\%m-\%d’, datetime(`\{admission\}’, `-1 day’)) AND strftime(`\%Y-\%m-\%d’, datetime(`\{chart\}’, `+1 day’)) AND \{condition\_value\};\\
 & SELECT * FROM Labevents JOIN d\_labitems ON Labevents.itemid = d\_labitems.itemid WHERE strftime(`\%Y-\%m-\%d’, chartevents.charttime) BETWEEN strftime(`\%Y-\%m-\%d’, datetime(`\{calculated\_time\}’, `-1 day’)) AND strftime(`\%Y-\%m-\%d’, datetime(`\{calculated\_time\}’, `+1 day’)) AND \{condition\_value\};\\
\SetCell[r=3]{c} Inputevents\_cv  & SELECT * FROM Inputevents\_cv JOIN d\_items ON Inputevents\_cv.itemid = d\_items.itemid WHERE strftime(`\%Y-\%m-\%d’, Inputevents\_cv.charttime) =`\{date\_value\}’ AND \{condition\_values\};\\
 & SELECT * FROM Inputevents\_cv JOIN d\_items ON Inputevents\_cv.itemid = d\_items.itemid WHERE strftime(`\%Y-\%m-\%d',Inputevents\_cv.charttime) BETWEEN strftime(`\%Y-\%m-\%d',datetime(`\{admission\}', `-1 day')) AND strftime(`\%Y-\%m-\%d',datetime(`\{admission\}', `+1 day')) AND \{condition\_value\};\\
 & SELECT * FROM Inputevents\_cv JOIN d\_items ON Inputevents\_cv.itemid = d\_items.itemid WHERE strftime(`\%Y-\%m-\%d',Inputevents\_cv.charttime) BETWEEN strftime(`\%Y-\%m-\%d', datetime(`\{admission\}', `-1 day')) AND strftime(`\%Y-\%m-\%d',datetime(`\{chart\}', `+1 day')) AND \{condition\_value\};\\
\SetCell[r=4]{c} Inputevents\_mv  & SELECT * FROM Inputevents\_mv JOIN d\_items ON Inputevents\_mv.itemid = d\_items.itemid WHERE strftime(`\%Y-\%m-\%d',`\{date\_value\}') BETWEEN strftime(`\%Y-\%m-\%d', Inputevents\_mv.starttime) AND strftime(`\%Y-\%m-\%d', Inputevents\_mv.endtime) AND \{condition\_value\};\\
&  SELECT * FROM Inputevents\_mv JOIN d\_items ON Inputevents\_mv.itemid = d\_items.itemid WHERE NOT (strftime(`\%Y-\%m-\%d', Inputevent\_mv.endtime) < datetime(`\{admission\}', '-1 day') OR strftime(`\%Y-\%m-\%d', Inputevents\_mv.starttime) > datetime (`\{admission\}', `+1 day')) AND \{condition\_value\};\\
&  SELECT * FROM Inputevents\_mv JOIN d\_items ON Inputevents\_mv.itemid = d\_items.itemid WHERE NOT (strftime(`\%Y-\%m-\%d',Inputevents\_mv.endtime) < datetime(`\{admission\}', `-1 day') OR strftime(`\%Y-\%m-\%d', Inputevents\_mv.starttime) > datetime (`\{chart\}', `+1 day'))AND \{condition\_value\};\\
&  SELECT * FROM Inputevents\_mv JOIN d\_items ON Inputevents\_mv.itemid = d\_items.itemid WHERE NOT (strftime(`\%Y-\%m-\%d',Inputevent\_mv.endtime) < datetime(`\{calculated\_time\}', `-1 day') OR strftime(`\%Y-\%m-\%d', Inputevents\_mv.starttime) > datetime (`\{calculated\_time\}', `+1 day'))AND \{condition\_value\};\\
\SetCell[r=3]{c} Microbiologyevents & SELECT * FROM Microbiologyevents JOIN D\_items AS spec\_items ON Microbiologyevents.spec\_itemid = spec\_items.itemid JOIN D\_items AS org\_items ON Microbiologyevents.org\_itemid = org\_items.itemid WHERE strftime(`\%Y-\%m-\%d', 
Microbiologyevents.charttime) = `\{date\_value\}' AND \{condition\_value\};\\
& SELECT * FROM Microbiologyevents JOIN D\_items AS spec\_items ON Microbiologyevents.spec\_itemid = spec\_items.itemid JOIN D\_items AS org\_items ON Microbiologyevents.org\_itemid = org\_items.itemid WHERE strftime(`\%Y-\%m-\%d', microbiologyevents.charttime) BETWEEN strftime(`\%Y-\%m-\%d', datetime(`\{admission\}', `-1 day')) AND strftime(`\%Y-\%m-\%d', datetime(`\{admission\}', `+1 day'))AND \{condition\_value\};\\
& SELECT *FROM Microbiologyevents JOIN D\_items AS spec\_items ON Microbiologyevents.spec\_itemid = spec\_items.itemid JOIN D\_items AS org\_items ON Microbiologyevents.org\_itemid = org\_items.itemid WHERE strftime(`\%Y-\%m-\%d', microbiologyevents.charttime) BETWEEN strftime(`\%Y-\%m-\%d', datetime(`\{admission\}', `-1 day')) AND strftime(`\%Y-\%m-\%d', datetime (`\{chart\}', `+1 day')) AND \{condition\_value\};\\
\SetCell[r=4]{c} Prescriptions & SELECT * FROM Prescriptions WHERE strftime(`\%Y-\%m-\%d', `\{date\_value\}') BETWEEN strftime(`\%Y-\%m-\%d', Prescriptions.startdate) AND strftime(`\%Y-\%m-\%d', Prescriptions.enddate) AND \{condition\_value\};\\
& SELECT * FROM Prescriptions WHERE NOT (strftime(`\%Y-\%m-\%d', Prescriptions.enddate) < datetime(`\{admission\}', `-1 day') OR strftime(`\%Y-\%m-\%d', Prescriptions.startdate) > datetime(`\{admission\}', `+1 day')) AND \{condition\_value\};\\
& SELECT * FROM Prescriptions WHERE NOT (strftime(`\%Y-\%m-\%d', Prescriptions.enddate) < datetime(`\{admission\}', `-1 day') 
OR strftime(`\%Y-\%m-\%d', Prescriptions.startdate) > datetime(`\{chart\}', `+1 day')) AND \{condition\_value\};\\
& SELECT * FROM Prescriptions WHERE NOT (strftime(`\%Y-\%m-\%d', Prescriptions.enddate) < datetime(`\{calculated\_time\}', `-1 day') OR strftime(`\%Y-\%m-\%d', Prescriptions.startdate) > datetime(`\{calculated\_time\}', `+1 day')) AND \{condition\_value\};\\
 \SetCell[r=3]{c} Outputevents & SELECT * FROM Outputevents JOIN d\_items ON Outputevents.itemid = d\_items.itemid WHERE strftime(`\%Y-\%m-\%d', Outputevents.charttime) = `\{date\_value\}' AND \{condition\_values\};\\
& SELECT * FROM Outputevents  JOIN d\_items ON Outputevents.itemid = d\_items.itemid WHERE strftime(`\%Y-\%m-\%d',Outputevents.charttime) BETWEEN strftime(`\%Y-\%m-\%d', datetime(`\{admission\}', `-1 day')) AND strftime(`\%Y-\%m-\%d', datetime(`\{admission\}', `+1 day'))AND \{condition\_value\};\\
& SELECT * FROM Outputevents JOIN d\_items ON Outputevents.itemid = d\_items.itemid WHERE strftime(`\%Y-\%m-\%d', Outputevents.charttime) BETWEEN strftime(`\%Y-\%m-\%d',datetime(`\{admission\}', `-1 day')) AND strftime(`\%Y-\%m-\%d', datetime(`\{chart\}', `+1 day'))AND \{condition\_value\};\\
\SetCell[r=1]{c} Procedures\_icd & SELECT * FROM Procedures\_icd JOIN D\_icd\_procedures ON Procedures\_icd.ICD9\_CODE=D\_icd\_procedures.ICD9\_CODE 
WHERE Procedures\_icd.hadm\_id={hadm\_id}AND (D\_icd\_procedures.LONG\_TITLE=`\{long\_title\}' OR D\_icd\_procedures.SHORT\_TITLE=`\{short\_title\}')\\
\SetCell[r=1]{c} Diagnoses\_icd & SELECT * FROM Diagnoses\_icd JOIN d\_icd\_diagnoses ON Diagnoses\_icd.ICD9\_CODE=d\_icd\_diagnoses.ICD9\_CODE WHERE Diagnoses\_icd.hadm\_id={hadm\_id} AND (d\_icd\_diagnoses.LONG\_TITLE=`\{long\_title\}' OR d\_icd\_diagnoses.SHORT\_TITLE=`\{short\_title\}')\\
\end{longtblr}

\newpage
\section{Author statement}
The authors of this paper bear all responsibility in case of violation of rights, etc. associated with \dname.

\end{document}